\documentclass[10pt, a4paper, logo, twocolumn, copyright]{googledeepmind} %

\usepackage{amsmath,amsfonts,bm}

\def\eqref#1{equation~\ref{#1}}

\def\1{\bm{1}}

\DeclareMathAlphabet{\mathsfit}{\encodingdefault}{\sfdefault}{m}{sl}
\SetMathAlphabet{\mathsfit}{bold}{\encodingdefault}{\sfdefault}{bx}{n}

\usepackage{hyperref}
\usepackage{url}
\usepackage{graphicx}
\usepackage{subcaption}
\usepackage{booktabs} %
\usepackage{wrapfig} %
\usepackage{enumitem}
\usepackage[htt]{hyphenat}
\usepackage{fontawesome}
\usepackage{makecell}
\usepackage{caption}
\usepackage{hyperref}

\usepackage{mathtools}
\usepackage{amsthm}
\usepackage[most]{tcolorbox}
\usepackage{moresize}
\usepackage{listings}
\usepackage{courier}
\usepackage{array}

\usepackage[capitalize,noabbrev]{cleveref}

\newcommand{\methodname}{Mind Evolution}
\newcommand{\refine}{Refinement through Critical Conversation}
\newcommand{\rcc}{RCC}
\newcommand{\sequential}{Sequential-Revision+}

\newcommand{\steg}{StegPoet}
\newif\ifsteg
\stegtrue

\usepackage[numbers]{natbib}
\title{Evolving Deeper LLM Thinking}

\author[a,b,1]{Kuang-Huei Lee}
\author[a,1]{Ian Fischer}
\author[c,2]{Yueh-Hua Wu}
\author[1]{Dave Marwood}
\author[1]{Shumeet Baluja}
\author[1,3]{Dale Schuurmans}
\author[1]{Xinyun Chen}

\affil[a]{First author contribution}
\affil[b]{Senior author contribution}
\affil[c]{Work done as a student researcher at Google DeepMind}
\affil[1]{Google DeepMind}
\affil[2]{UC San Diego}
\affil[3]{University of Alberta}

\begin{abstract}

We explore an evolutionary search strategy for scaling inference time compute 
in Large Language Models.
The proposed approach, \methodname, uses a language model to generate,
recombine and refine candidate responses.
The proposed approach avoids the need to formalize the underlying inference
problem whenever a solution evaluator is available.
Controlling for inference cost, we find that \methodname\ significantly
outperforms other inference strategies such as Best-of-N and Sequential Revision
in natural language planning tasks.
In the TravelPlanner and Natural Plan benchmarks, \methodname\ solves more
than 98\% of the problem instances using Gemini~1.5~Pro
without the use of a formal solver.

\end{abstract}

\begin{document}

\maketitle

\section{Introduction}
\label{sec:intro}

How can a large language model (LLM) be guided to \emph{think deeper} about
a complex problem and leverage inference time compute to improve its problem
solving ability?
Prior research has investigated various strategies for leveraging
inference time compute,
such as chain-of-thought~\cite{wei2022chain,kojima2022large},
self-consistency~\cite{wang2023selfconsistency}, 
sequential revision based on feedback~\cite{shinn2024reflexion,madaan2024self,chen2024teaching,kim2023rci,bai2022constitutional},
and search guided by auxiliary verifiers or evaluators~\cite{yao2023tree}.
When a solution evaluator is available, 
search strategies have 
an advantage of being able to reliably improve problem solving ability with increased compute.
For example, methods
such as Best-of-N~\cite{brown2024large,liang2024improving,lightman2023let}
and tree search~\cite{snell2024scaling} naturally exploit additional compute
to explore a larger set of solution candidates,
thereby increasing the probability of finding a successful solution.

\begin{figure*}
    \centering
    \includegraphics[width=\linewidth]{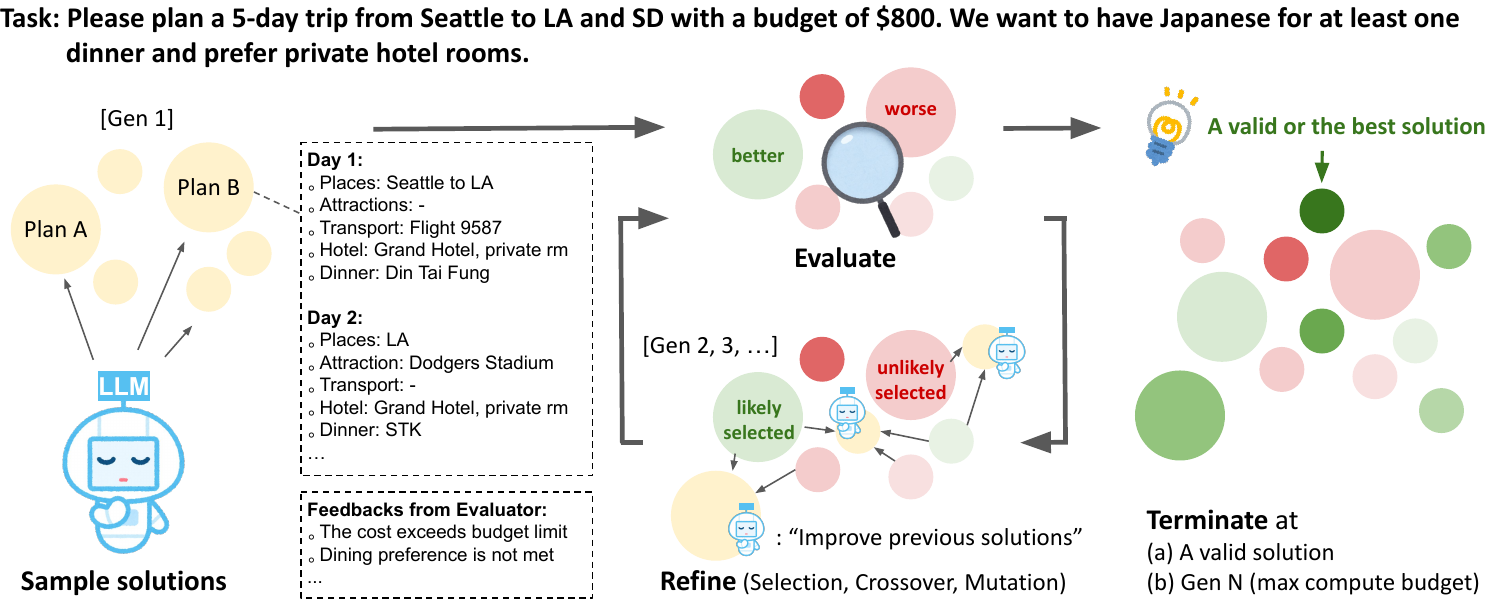}
    \caption{%
\methodname\ is a genetic-based evolutionary search strategy that operates
in natural language space. 
The figure illustrates how \methodname\ evolves a population of solution
candidates toward higher quality candidates for a travel planning task.
The candidate population is improved through an iterative process,
where an LLM is used to recombine and refine candidates in each iteration.
    }
    \label{fig:overview}
\end{figure*}

To better exploit inference time compute,
we propose an evolutionary search strategy for LLMs that combines free-flowing
stochastic exploration with large-scale iterative refinement.
We refer to this approach as \emph{\methodname}.
As illustrated in \Cref{fig:overview}, \methodname\ is a genetic search
strategy that evolves a diverse population of candidate solutions,
leveraging an LLM to generate, recombine and refine solution
candidates based on feedback from an evaluator.
The overall process is analogous to combining divergent thinking 
(free-flowing parallel idea exploration) with convergent thinking 
(idea evaluation and selection), 
considered as hallmarks of intelligent problem solving
behavior~\cite{guilford1967nature}. 

Unlike Best-of-N, which searches broadly by generating independent candidates
for evaluation,
\methodname\ searches both broadly and deeply, exploring a diverse set of
candidates and refining the most promising alternatives.
Unlike sequential reasoning approaches, 
such as self-refinement or tree search~\cite{snell2024scaling,lightman2023let},
which require evaluation of individual reasoning steps,
\methodname\ performs global refinement of complete solutions,
and therefore only requires a global solution evaluator
rather than a stepwise process reward.
Also, typical of evolutionary methods, \methodname\ can be easily parallelized.

There has been prior work on combining evolutionary search with LLMs,
primarily in the literature on evolutionary program generation~\cite{romera2024mathematical,hemberg2024evolving,liventsev2023fully,lehman2023evolution,chen2023evoprompting}.
However, this prior work focuses on searching through formal program spaces,
using guidance from execution feedback or code explanation.
By contrast, \methodname\ is not restricted to searching in a formal space.
This allows \methodname\ to be applied to problems that
are not formalized, or remain difficult to formalize,
as long as a programmatic solution evaluator is available.
In particular, we focus on natural language planning tasks 
where candidate solutions can still be automatically parsed, evaluated
and critiqued using an implementable oracle evaluator.
This approach exploits the observation that it is often easier to evaluate
the quality of a candidate solution than it is to generate good solutions
for a given problem~\cite{gareyjohnson79}.

In the domain of natural language planning,
we consider the TravelPlanner~\cite{xie2024travelplanner} and 
Natural Plan~\cite{zheng2024natural} benchmarks,
where constraint satisfaction problems are expressed in natural language
without any explicit formalization of the underlying
objectives, constraints or variables.
These problems require a set of interconnected decisions
that satisfy a set of global and local constraints.
For example, in TravelPlanner, a travel plan should be produced
that respects various accommodation and dinning constraints, 
while also considering budget limitations and other preferences,
all expressed solely in natural language.
To date, LLMs have yet to achieve good performance on these tasks
without the aid of formal solvers~\citep{hao2024large}.
For example, Gemini 1.5 Flash and o1-preview only achieve a success rate of
5.6\% and 11.7\% on TravelPlanner respectively,
while for the Meeting Planning domain in Natural Plan,
they respectively only achieve 20.8\% and 44.2\%.
Even exploiting Best-of-N over 800 independently generated responses,
Gemini 1.5 Flash still only achieves 55.6\% success on
TravelPlanner and 69.4\% on Meeting Planning.
In this paper, we show that exploration and refinement with evolutionary search
can notably improve problem solving ability.
In particular, when controlling for inference time compute, 
\methodname\ allows Gemini 1.5 Flash to achieve a 95.6\% success rate on
TravelPlanner and 85.0\% on Meeting Planning.
We further experiment with a two-stage approach, 
where any unsolved problem instances are subsequently tackled by
\methodname\ with Gemini 1.5 Pro,
which leads to 100\% success on TravelPlanner and 98.4\% on Meeting Planning.
All of the experiments in this paper only use off-the-shelf LLMs
without any finetuning.

To our knowledge, the only prior work that achieves comparable performance on
the TravelPlanner benchmark is~\citep{hao2024large},
which leverages an auxiliary formal solver and requires the LLM to
first translate a given problem instance into an equivalent formalization.
In general, it takes significant effort and expertise to correctly
formalize a problem expressed in natural language;
prompting an LLM to correctly perform such a translation
requires at least as much domain expertise.
\methodname\ removes this constraint by directly optimizing solutions in the space of natural language.

\ifsteg
Finally, we introduce a new benchmark problem, \steg,
that involves encoding a hidden message in a generated essay, story or poem.
This form of stenography~\cite{provos2003hide} is difficult to formalize and solve,
yet a hidden message detector can still be implemented to programmatically
guide the search.
Our motivation is to demonstrate the applicability of 
search beyond natural language domains that can be easily formalized.
We find that \methodname\ allows Gemini 1.5 Pro to
achieve a success rate of 87\% in this task.
\fi

\section{Related Work}
\label{sec:related_work}

\paragraph{Pairing LLMs with Evolutionary Search}
In addition to the program generation studies discussed in \Cref{sec:intro}, several recent works have explored combining LLMs and evolution for numerical optimization~\citep{liu2023large,brahmachary2024large} and combinatorial optimization~\citep{liu2024large,ye2024large}.
The problem spaces we tackle in this work, such as natural language planning, can also be viewed as combinatorial optimization problems -- optimizing plans subject to constraints specified in natural language.
In contrast to these previous studies, we focus on evolving solutions in natural language spaces instead of formal spaces. 
This removes the requirement of task formalization, which requires significant effort and expert knowledge for each task instance.

Other works have also applied evolutionary search to prompt optimization, with the goal of improving performance on target tasks~\citep{yuan2024evoagent, fernando2023promptbreeder, guo2023connecting}.
Among these, EvoAgent~\citep{yuan2024evoagent} also evaluated their approach on the TravelPlanner benchmark. 
In contrast to our work, which performs evolutionary search directly on plans, EvoAgent evolves new LLM agents to form a multi-agent system for problem solving. Their best success rate on the TravelPlanner validation set was $7.2\%$ with GPT-4, while our approach achieved over $95\%$ with Gemini 1.5 Flash.

\paragraph{Pairing LLMs with Evaluators}
In this work, we evaluate solutions with program-based evaluators during the evolutionary search.
The idea of integrating execution-based evaluators in the inference loop has been widely adopted in the literature of code generation, where the execution environment provides feedback for the LLM to fix bugs in the generated code~\cite{chen2023codet,le2022coderl,liu2023rltf,zhang2023algo,chen2024teaching,hemberg2024evolving,liventsev2023fully,lehman2023evolution,chen2023evoprompting,shinn2024reflexion}.

Other prior work has also considered using learned verifiers, reward models, or self-evaluation for response refinement~\citep{kirchner2024prover,madaan2024self}, search~\citep{snell2024scaling,brown2024large,cobbe2021training,yao2023tree,setlur2024rewarding}, and improving model learning~\citep{wang2024multi,lightman2023let,park2023generative,bai2022constitutional}.
These approaches can often be applied to wider domains and free-form solutions, 
but learned feedback models or self-evaluators can be noisy and are not perfectly reliable.
We leave consideration of such approximate feedback mechanisms for future work.

\section{Method}
\label{sec:method}

\methodname\ employs a genetic search strategy,
combined with an LLM and a tailored set of prompts,
to orchestrate an efficient search for solutions to
natural language planning tasks.
Before describing \methodname\ in detail,
we first provide a brief overview of language-based genetic algorithms.

\subsection{Language-based Genetic Algorithm Overview}
\label{subsec:lbga}

Genetic algorithms~\citep{holland1975, golberg1989genetic, mitchell1998introduction} are a meta-heuristic inspired by natural selection.
In a genetic algorithm, a population of candidate solutions is evolved
toward populations that contain a greater proportion of higher quality
individuals with respect to a target optimization objective.
Such an objective is also often referred to as the ``fitness'' function.
Each individual candidate has a genetic representation that can be
mutated and recombined with others.

Evolutionary search usually begins with a population of independently generated
candidate solutions.
In each generation, the fitness of every individual is evaluated 
with respect to the target objective.
Candidates are then stochastically selected for reproduction 
based on their fitness (``selection'').
In reproduction, the genetic representations of selected parents
are combined (``crossover'') and potentially altered (``mutation'')
to produce new child solutions.
Such a process creates the next generation of children,
which then enter the population.
Population fitness generally increases over successive generations,
as parents with greater fitness are more likely to be selected for
recombination.

\paragraph{Island Model}
To sustain diversity in an evolving population it is also helpful to introduce
an island model~\citep{tanese1989distributed, cantu1998survey},
where distinct sub-populations (``islands'') are created and evolved
independently between ``migration'' and ``island reset'' events
that occur at specified frequencies.
For a migration operation, the solutions on one island are stochastically 
chosen based on fitness to migrate to an adjacent island.
For an Island Reset operation, the populations on islands with low overall
fitness are replaced by strong solutions from the global population,
which also has a selection effect.
The island model has been adopted in recent successful efforts,
such as FunSearch~\citep{romera2024mathematical}.

\paragraph{Language-based Genetic Representation}
The individual candidates in a language-based genetic algorithm are
represented by natural language.
This allows the strong language understanding and generation capabilities of 
an LLM to be leveraged to implement powerful recombination
(crossover and mutation) and island reset operations through prompting.

\subsection{\methodname}
\label{subsec:our_method}

\begin{table*}%
    \small
    \centering
    \begin{tabular}{r|c|l}
        Parameter & Default Value & Description \\
        \hline
        $N_{\text{gens}}$ & 10 & The maximum number of generations to search for a solution. \\
        $N_{\text{island}}$ & 4 & How many independent populations to evolve. \\
        $N_{\text{convs}}$ & 5 & How many conversations per island. \\
        $N_{\text{seq}}$ & 4 & How many turns per conversation. \\
        \hline
        $N_{\text{reset interval}}$ & 3 & How frequently to reset islands in generations. \\
        $N_{\text{reset}}$ & 2 & How many islands to reset. Lowest mean score islands are chosen. \\
        $N_{\text{top}}$ & 5 & How many starting parents to transfer to islands when reset. \\
        $N_{\text{candidate}}$ & 15 & How many candidate parents to consider when resetting islands with the LLM. \\
        \hline
        $N_{\text{parent}}$ & 5 & Maximum number of parents a conversation can have. \\
        $Pr_{\text{no parents}}$ & 1/6 & Probability of a conversation having no parents. \\
        $N_{\text{emigrate}}$ & 5 & How many plans to emigrate to the next island after each island. \\
        $N_{\text{retries}}$ & 5 & How many times to try to generate a plan before giving up at each turn. \\
    \end{tabular}
    \caption{%
        Definition of hyperparameters in \methodname. 
        Unless otherwise specified, the experiments in work use the default values.
        The product of the first four hyperparameters gives the maximum number of candidate solutions generated (800 in the default setting).
    }
    \label{tab:hyperparameters}
\end{table*}

\Cref{fig:overview} illustrates the design of \methodname,
with its hyperparameters listed in \cref{tab:hyperparameters}.
The core components of \methodname\ are:
\begin{enumerate}
\item
the specific choices for the selection and migration operations;
\item
the set of prompts that implement the initialization, recombination 
(crossover and mutation), and island reset operations with an LLM;
\item
the fitness function that evaluates the quality of a given solution 
and optionally provides feedback on issues detected.
\end{enumerate}
The overall evolution process is repeated until a valid solution is found,
or until $N_{\text{gens}}$ generations have been completed,
after which the best scoring candidate is returned.

\paragraph{Fitness Evaluation}
As discussed in \Cref{sec:intro},
we implement a fitness function for each problem domain,
where candidate solutions are parsed and evaluated programmatically.
In principle, any function that can evaluate solution quality can be used,
including LLM evaluation.
The evaluation function plays three key roles in \methodname:
(1) scoring solutions by measuring the optimization objective, if any;
(2) verifying whether the solution satisfies given constraints;
and (3) providing corresponding textual feedback.
For example, the evaluation function for the Meeting Planning task scores a
proposed plan and provides textual feedback based on how many constraints are
violated
(e.g. meetings conflict with existing schedules),
how many valid meeting events are included in the schedule,
and whether the plan follows the required format
(see \cref{subsec:eval_func} for more details).
We have found that using textual feedback is important empirically,
as shown in our ablation study in \cref{subsec:ablations}.

Note that for many classical search problems (e.g., NP-complete problems),
verifying solutions can be much easier than solving the problem~\cite{gareyjohnson79}.
Similarly, we observe that it is possible to write an evaluation
function for the natural language planning tasks we consider.
The ability to check the correctness of a candidate solution does not
obviously lead to the ability to generate a valid solution in the tasks we consider.
That is,
implementing an evaluation function is not equivalent to solving the task.

\paragraph{Population Initialization}
Given a target problem, we independently sample $N_{\text{convs}}$ initial
solutions by prompting an LLM with a description of the problem,
any information needed for solving the problem, and relevant instructions.
If $N_{\text{seq}}>1$, 
each of these initial solutions is then evaluated and refined sequentially 
through $N_{\text{seq}}-1$ additional turns of the 
``\refine'' process explained below.
In total, this initialization procedure generates
$N_{\text{convs}} \times N_{\text{seq}}$ candidate solutions,
which forms the initial population on the first island for the first generation.

\begin{figure}[t]
    \centering
    \includegraphics[width=0.48\textwidth]{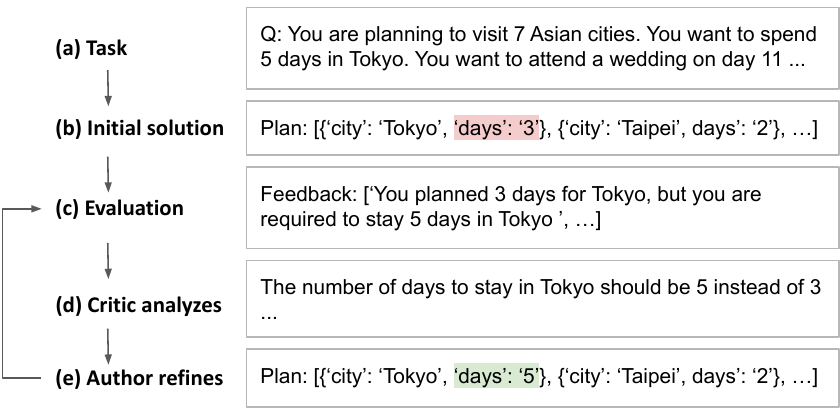}
    \caption{%
Illustrating the \refine\ (\rcc) process,
where an initial solution is first proposed, then evaluated and subjected to 
feedback from a critic, after which an author proposed a refined solution
and the process iterates.
    }
    \label{fig:critical_thinking}
\end{figure}

\paragraph{\refine\ (\rcc)}
Given a candidate solution
(or a set of candidate solutions for the process of recombination)
we leverage an LLM to generate an improved solution
by organizing a critical conversation between
a ``critic'' character and an ``author'' character,
as illustrated in \cref{fig:critical_thinking}.
Separating these two roles is intended to improve the critical thinking
ability of an LLM.
Each conversational turn is structured as a prompt-driven
process, where solutions are refined based on critical feedback,
similar to Reflexion~\cite{shinn2024reflexion}.
In particular,
the critic first analyzes the candidate solution(s) provided as input,
interprets the textual evaluation feedback,
and suggest ways to correct any issues presented in the feedback.
The author then proposes a single refined solution based on the input
candidate(s), the subsequent evaluation(s), and the critic's analyses.
The specific prompts used to drive these conversations
are given in \cref{subsec:prompt_design}.
An ablation study in \Cref{subsec:ablations} shows that the critic’s analysis
step provides substantial performance improvements.

\paragraph{Selection}
To produce the next generation of an island, 
we follow Boltzmann tournament selection~\cite{goldberg1990note}
where $0$ to $N_{\text{parent}}$ parents are stochastically sampled from the
population according a probability distribution
that is derived from a softmax transformation of their fitness scores.
In this way, higher-performing solutions are more likely to be selected for
reproduction, while other candidates can still be occasionally selected for
diversity.

\paragraph{Crossover and Mutation}
We implement the crossover and mutation operations as a single recombination step,
where an LLM is instructed to improve a given set of parents
using the \rcc\ process described above (\Cref{fig:critical_thinking}).
In particular, for recombination we sample $1$ to $N_{\text{parent}}$ parents
and alter Step (b) in \Cref{fig:critical_thinking}
to first incorporate the evaluation results of the parents, then apply the critic to all parents
and propose the revised solution as an ``initial solution'' for the next generation.
Then, if $N_{\text{seq}}>1$, we continue to follow Steps (c)(d)(e) to sequentially generate
$N_{\text{seq}}-1$ child solutions by refining each previous child using the \rcc\ process.

For each generation on each island, $N_{\text{convs}} \times N_{\text{seq}}$
child solutions are added to the island population,
with duplicate solutions removed.
For selection, we follow a Boltzmann tournament instead of explicitly retiring
candidate solutions, except when performing an Island Reset below.

\paragraph{Migration between Islands}
Between migration events, each island population is evolved independently.
During a migration,
the top $N_{\text{emigrate}}$ solutions are cloned from the current Island $i$
to the next Island $i+1$ after completing the generation on the current island
(we update the populations on the islands sequentially from $1$ to
$N_{\text{island}}$).
Migration is performed cyclically between the islands,
so emigrants from Island $N_{\text{island}}$ arrive at Island $1$.
We have found that this form of cyclic migration accelerates the overall
evolution process.

\paragraph{Island Reset}
Island reset happens every $N_{\text{reset interval}}$ generations.
During an Island Reset event, the top performers are first selected from the
global population, the populations on $N_{\text{reset}}$ islands with the
lowest average scores are retired, and the selected top performers are cloned
onto the reset islands.
To select top performers, we explore two approaches:
(1) directly select the top $N_{\text{top}}$ candidates according to fitness;
and
(2) first select the top $N_{\text{candidate}}$ candidates according to fitness,
then prompt the LLM to select $N_{\text{top}}$ good candidates from this pool
that are substantially different from each other.
The ablation study in \cref{subsec:ablations} show that the latter strategy,
using an LLM for Island Reset, achieves better performance.

\section{Experiments}
\label{sec:experiments}

\paragraph{Tasks}
We evaluate \methodname\ on three benchmark
natural language planning domains:
two tasks from Natural Plan \citep{zheng2024natural},
including Trip Planning (\Cref{subsec:np_trip_planning}) 
and Meeting Planning (\Cref{subsec:np_meeting_planning}),
and the TravelPlanner~\citep{xie2024travelplanner} benchmark
(\Cref{subsec:travel_planner}).
(We omit the Calendar Scheduling task from Natural Plan,
since these problems can be solved by enumeration.)
Implementation details for each task is provided in 
\cref{sec:implementation_details}, 
including the prompts (\cref{subsec:prompt_design})
and evaluation functions used (\cref{subsec:eval_func}).

\paragraph{Models}
We use Gemini 1.5 Flash (gemini-1.5-flash-001) as the default LLM
in our experiments below.
The hyperparameters used when applying \methodname\ to Flash
are specified in \cref{tab:hyperparameters}.
In addition to evaluating \methodname\ with the Flash model,
we also investigate a two-stage approach,
where Gemini 1.5 Pro model (gemini-1.5-pro-exp-0827)
is used to tackle problems that are not solved
within the $N_{\text{gens}}$ generation limit.
Such a two-stage approach provides better cost-efficiency than using the 
Pro model on every problem instance.
When applying \methodname\ to the Pro model we alter the hyperparameters
from those specified in \cref{tab:hyperparameters} to:
$N_{\text{convs}}=8$, $N_{\text{seq}}=3$, $N_{\text{parent}}=10$,
$Pr_{\text{no parents}}=1/5$.

\paragraph{Baselines}
For each task, we compare \methodname\ to three baseline search strategies
that use the same solution evaluator and task-specific prompts:
\begin{enumerate}
\item
\textbf{1-Pass},
where a solution is proposed using a single forward pass of the LLM.
\item
\textbf{Best-of-N}~\cite{brown2024large},
where up to 800 candidate solutions are independently generated until a
successful solution is found
(the same upper bound as \methodname).
\item
\textbf{\sequential},
where 10 candidate solutions are proposed independently,
then revised separately for 80 turns using the
\rcc\ process (\cref{fig:critical_thinking}).
Note that 10 independent threads of 80-turn refinements are used instead of
a single 800-turn refinement, because we rarely observe improvements after 80
turns.
This baseline is similar to running 10 trials of multi-turn
Reflexion~\citep{shinn2024reflexion}.
\end{enumerate}
Additionally, for reference,
we also include an additional 1-Pass baseline that uses OpenAI o1-preview.

\paragraph{Metrics}
We measure Success Rate as the percentage of problem instances
that are solved completely within a benchmark domain,
separating the validation and test sets.
(Note that the Success rate is referred to as Solve Rate in 
Natural Plan~\cite{zheng2024natural} 
and Final Pass Rate in TravelPlanner~\cite{xie2024travelplanner}.)

To assess the cost of inference compute we report
the number of LLM calls, 
the number of input and output tokens,
and the total API cost of calling the LLM.
(These costs are given in US Dollars, using prices from October 2024 when the
experiments were conducted. 
The base rates are listed in \cref{sec:pricing}.)
Note that assessing computational cost is particularly important when evaluating
search strategies like \methodname, since search is more expensive than 
generating a single solution.
These statistics can help researchers and developers understand the cost-benefit trade-offs when using search to enhance LLM problem solving ability.

\begin{table*}[!t]
    \setlength{\tabcolsep}{0.4em}
    \small
    \centering
    \begin{tabular}{|r|c|r|r|r|r|r|}
        \hline
         & Set & Success Rate & LLM Calls & Input Tokens & Output Tokens & API Cost (Oct 2024) \\
        \hline
        \multicolumn{7}{|l|}{\textbf{TravelPlanner~\cite{xie2024travelplanner}}} \\
        \hline
        1-Pass & val & $10/180=5.6\%$ & 1 & $0.009$M & $0.001$M & US\$$0.001$ \\
        (o1-preview 1-Pass) & val & $21/180=11.7\%$ & 1 & $0.008$M & $0.008$M & US\$$0.601$\\
        Best-of-N & val & $100/180=55.6\%$ & 472 & $4.44$M & $0.47$M & US\$$0.47$\\
        \sequential & val & $149/180=82.8\%$ & 280 & $35.53$M & $0.29$M & US\$$2.75$\\
        \bf{\methodname} & val & $172/180=$ $\bf{95.6\%}$ & 174 & $3.10$M & $0.18$M & US\$$0.29$\\
        \bf{(+pro)} & val  & $180/180=$ $\bf{100\%}$ & (257) & ($3.25$M) & ($0.19$M) & (US\$$0.54$)\\
\hline
        \bf{\methodname} & test & $952/1000=$ $\bf{95.2\%}$ & 167 & $3.02$M & $0.18$M & US\$$0.28$\\
        \bf{(+pro)} & test & $999/1000=$ $\bf{99.9\%}$ & (67) & ($3.05$M) & ($0.18$M) & (US\$$0.33$)\\
         
        \hline
        \multicolumn{7}{|l|}{\textbf{Natural Plan~\cite{zheng2024natural} Trip Planning}} \\
        \hline
        1-Pass & val & $66/320=20.6\%$ & 1 & $0.002$M & $0.001$M & $<$US\$$0.001$\\
        (o1-preview 1-Pass) & val & $116/320=36.2\%$ & 1 & $0.002$M & $0.008$M & US\$$0.53$\\
        Best-of-N & val & $247/320=77.2\%$ & 274 & $0.61$M & $0.18$M & US\$$0.10$\\
        \sequential & val & $238/320=74.4\%$ & 391 & $41.57$M & $0.38$M & US\$$3.23$\\
        \bf{\methodname} & val & $308/320=\textbf{96.2\%}$ & 168 & $1.48$M & $0.19$M & US\$$0.17$\\
        \bf{(+pro)} & val & $320/320=\textbf{100\%}$ & (111) & ($1.51$M) & ($0.19$M) & (US\$$0.22$)\\
\hline
        \bf{\methodname} & test & $1204/1280=\textbf{94.1\%}$ & 196 & $1.78$M & $0.22$M & US\$$0.20$\\
        \bf{(+pro)} & test & $1275/1280=\textbf{99.6\%}$ & (211) & ($1.86$M) & ($0.24$M) & (US\$$0.37$)\\

        \hline
        \multicolumn{7}{|l|}{\textbf{Natural Plan~\cite{zheng2024natural} Meeting Planning}} \\
        \hline
        1-Pass & val & $104/500=20.8\%$ & 1 & $0.007$M & $0.001$M & US\$$0.001$\\
        (o1-preview 1-Pass) & val & $221/500=44.2\%$ & 1 & $0.006$M & $0.006$M & US\$$0.47$\\
        Best-of-N & val & $347/500=69.4\%$ & 444 & $3.99$M & $0.31$M & US\$$0.39$\\
        \sequential & val & $310/500=62.0\%$ & 484 & $32.16$M & $0.40$M & US\$$2.53$\\
        \bf{\methodname} & val & $425/500=\textbf{85.0\%}$ & 406 & $5.35$M & $0.41$M & US\$$0.52$\\
        \bf{(+pro)} & val & $492/500=\textbf{98.4\%}$ & (890) & ($13.36$M) & ($0.91$M) & (US\$$2.55$)\\
\hline
        \bf{\methodname} & test & $419/500=\textbf{83.8\%}$ & 394 & $5.24$M & $0.40$M & US\$$0.51$\\
        \bf{(+pro)} & test & $491/500=\textbf{98.2\%}$ & (828) & ($12.25$M) & ($0.83$M) & (US\$$2.34$)\\
        \hline
    \end{tabular}
    \caption{%
        Experimental results on benchmark natural language planning tasks.
        ``(+pro)'' denotes the two-stage results, where we use Gemini 1.5 Pro to solve the problems that were not solved in experiments using Gemini 1.5 Flash. 
        Number of LLM calls, token counts, and API cost are averaged across the validation or test problem set, and they are calculated only on the remaining problems for the ``(+pro)'' experiments.
        Here, we also show OpenAI o1-preview results as a reference.
    }
    \label{tab:planning_results}
\end{table*}

\subsection{TravelPlanner}
\label{subsec:travel_planner}

TravelPlanner~\cite{xie2024travelplanner} is a natural language planning
benchmark that simulates the problem of organizing a trip plan for a user
who expresses preferences and constraints.
We focus on the sole-planning mode
(see \cite{xie2024travelplanner} for details),
where each problem instance consists of a list of options regarding
accommodation, restaurants, attractions and transportation,
plus additional constraints that specify user preferences for budget, cuisine,
etc.
A plan is evaluated based on whether it satisfies the user preferences and
commonsense constraints.

\Cref{tab:planning_results} gives detailed results that compare
the overall Success Rate and computational cost of \methodname\ 
versus the baseline strategies. 
In terms of Success Rate, \methodname\ clearly outperforms the baseline
strategies, achieving over 95\%.
By comparison, \sequential\ provides a reasonable baseline, achieving
almost 83\%, while Best-of-N struggles, achieving only 55.6\%.
Overall,
these results demonstrate a clear advantage of an evolutionary strategy that
combines a broad search, through stochastic exploration, with a deep
search that leverages an LLM for solution refinement.

Considering the two-stage approach, where \methodname\ uses
Gemini 1.5 Pro for any unsolved problems,
we find that nearly the entire dataset can be solved,
achieving a 100\% success rate on validation and
99.9\% on test problems respectively.
The only work we are aware of that comes close to this success rate is
\cite{hao2024large}, which uses GPT-4 for auto-formalization then leverages
a formal solver to achieve 98.9\% and 97.0\% on validation and test
respectively.
\methodname\ achieves comparable results without requiring a formal solver.

\begin{figure}[t]
    \centering
    \includegraphics[width=0.48\textwidth]{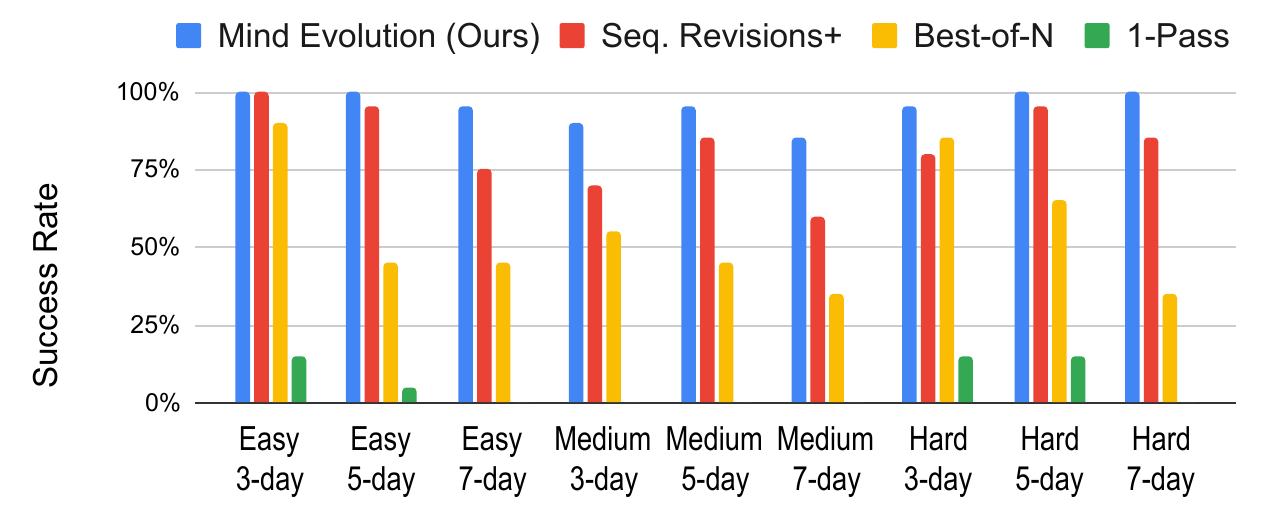}
    \caption{%
Success rate on the validation set of the TravelPlanner benchmark,
organized by problem instance difficulty and the number of travel days.
    }
    \label{fig:hist_travelplanner}
\end{figure}

Finally, we note that the TravelPlanner dataset is organized into
three levels of difficulty (Easy, Medium, Hard) and three trip durations
(3 days, 5 days, 7 days), rendering 9 different problem classes. 
\cref{fig:hist_travelplanner} presents a breakdown of the success rates
achieved across these different categories,
showing that the success rates of 1-Pass and Best-of-N decline when planning
for more travel days, but the trend is less clear for \methodname\ and
\sequential, both of which iteratively refine proposed solutions.

\subsection{Natural Plan -- Trip Planning}
\label{subsec:np_trip_planning}

The Trip Planning task \citep{zheng2024natural}
involves finding an itinerary that consists of a sequence of cities to visit
and number of days in each that satisfies flight connectivity
and scheduling constraints --
see \Cref{tab:trip_planning_example} for a problem instance.
We split the benchmark into 320 validation and 1,280 test instances
(described in more detail in \cref{sec:data_splits}).

The results in \cref{tab:planning_results} again show that \methodname\ 
strongly outperforms the baselines on this task,
achieving 96.2\% on the validation and 94.1\% on the test instances.
\cref{tab:planning_results} also shows a qualitative comparison between the 
results produced by \methodname\ and the baseline strategies.
Note that Best-of-N performs better in this scenario (77.2\%), even beating 
\sequential\ (74.4\%).
We find that for the two-stage approach,
\methodname\ achieves 100\% on the validation set and 99.6\% on the test set.
These findings again highlight the benefit of evolutionary search
versus simple sampling and sequential refinement.

\begin{figure}[t]
    \centering
    \includegraphics[width=0.48\textwidth]{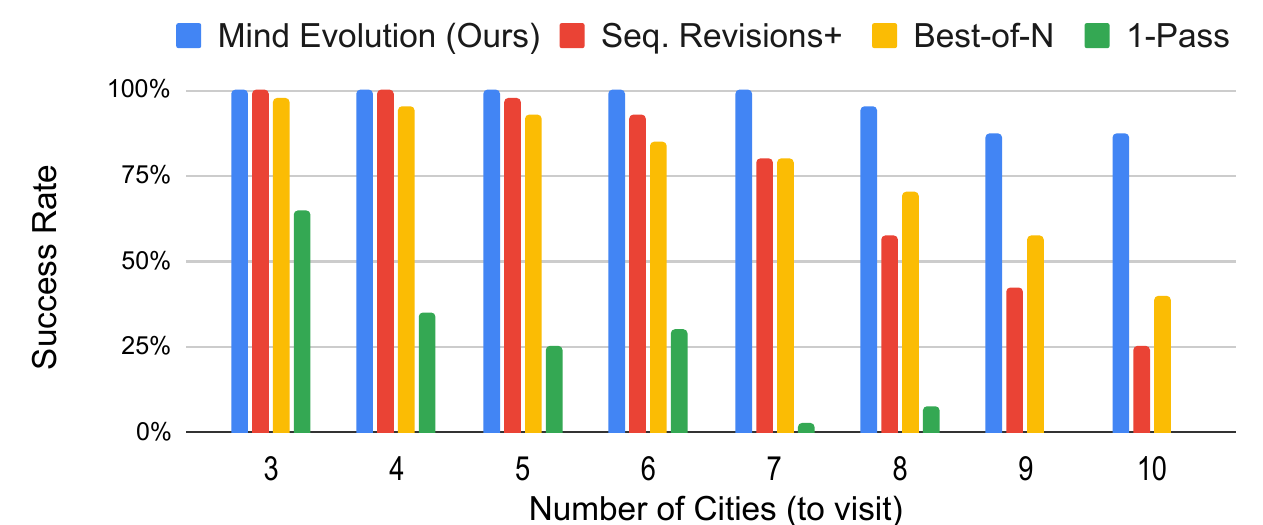}
    \caption{Success rate on the validation set of the Trip Planning benchmark per number of cities to visit.}
    \label{fig:hist_trip_planning}
\end{figure}

Finally,
we note that the difficulty of this task varies with the number of cities to
visit, ranging from 3 to 10.
\cref{fig:hist_trip_planning} shows a breakdown of the Success Rate in terms 
of number of cities, where the relative advantage of \methodname\ appears to
increase as the number of cities grows.

\begin{table*}[t]
    \setlength{\tabcolsep}{0.4em}
    \small
    \centering
    \arrayrulecolor{gray!70}
    \begin{tabular}{|lp{0.77\textwidth}|}
    \hline
    \multicolumn{2}{|p{0.95\textwidth}|}{
{\fontfamily{cmss}\selectfont
Q: You plan to visit 5 European cities for 16 days in total. You only take direct flights to commute between cities. You want to spend 5 days in Madrid. From day 3 to day 7, there is a annual show you want to attend in Madrid. You plan to stay in Zurich for 3 days. You would like to visit Frankfurt for 3 days. You would like to visit Santorini for 6 days. You are going to attend a wedding in Santorini between day 7 and day 12. You want to spend 3 days in Riga.

\vspace{+0.3em}

Here are the cities that have direct flights:
Zurich and Riga, Frankfurt and Riga, Santorini and Zurich, Madrid and Zurich, Frankfurt and Zurich, Madrid and Santorini, Frankfurt and Madrid.

\vspace{+0.3em}

Find a trip plan of visiting the cities for 16 days by taking direct flights to commute between them.}} \\
& \\
\textbf{Method} & \textbf{Answer} \\
1-Pass &
{\fontfamily{cmss}\selectfont
Madrid \colorbox{pink}{(Day 1-7)} \faAngleRight\ Santorini (Day 7-12) \faAngleRight\ Zurich (Day 12-14) \faAngleRight\ Frankfurt (Day 14-16) \faAngleRight\ Riga \colorbox{pink}{(Day 16-19)} \textcolor{red}{\faClose\ 7 days for Madrid instead of 5; 4 days for Riga instead of 3; 19 days in total instead of 16.}
} \\
Best-of-N &
{\fontfamily{cmss}\selectfont
Madrid \colorbox{pink}{(Day 1-7)} \faAngleRight\ Santorini (Day 7-12) \faAngleRight\ Zurich (Day 12-14) \faAngleRight\ Frankfurt (Day 14-16) \faAngleRight\ Riga \colorbox{pink}{(Day 16-16)} \textcolor{red}{\faClose\ 7 days for Madrid instead of 5; 1 day for Riga instead of 3.}
} \\
Sequential Revisions+ &
{\fontfamily{cmss}\selectfont
Zurich (Day 1-3) \faAngleRight\ Frankfurt (Day 3-5) \faAngleRight\ Riga (Day 5-7) \colorbox{pink}{\faAngleRight} Santorini (Day 7-12) \faAngleRight\ Madrid \colorbox{pink}{(Day 12-16)} \textcolor{red}{\faClose\ omitted the show in Madrid (Day 3-7); no direct flight from Riga to Santorini.}
} \\
Mind Evolution (ours) &
{\fontfamily{cmss}\selectfont
Frankfurt (Day 1-3) \faAngleRight\ Madrid (Day 3-7) \faAngleRight\ Santorini (Day 7-12) \faAngleRight\ Zurich (Day 12-14) \faAngleRight\ Riga (Day 14-16) \textcolor{green}{\faCheck}
} \\

    \hline         
    \end{tabular}
    \caption{
An example problem instance from the Trip Planning task in Natural Plan,
with the predicted plans from Mind Evolution and the baselines.
1-Pass and Best-of-N both make mistakes on number of days to stay,
but satisfy the requirements of being in Madrid and Santorini on specific days.
The \sequential\ plan omits the annual show in Madrid and plans a non-existent
flight, but is correct in the number of days.
In contrast, the \methodname\ plan satisfies all specified requirements.}
    \label{tab:trip_planning_example}
\end{table*}

\subsection{Natural Plan -- Meeting Planning}
\label{subsec:np_meeting_planning}

For the Meeting Planning task a sequence of meetings should be scheduled
to maximize the number of meetings between individuals subject to
availability, location and travel time constraints \cite{zheng2024natural}.
This task differs from TravelPlanner and Trip Planning
in that not every meeting can be scheduled for every problem instance,
implying that it is not possible to know whether an optimal solution
has been reached.
Therefore,
to obtain the results shown in \cref{tab:planning_results},
we allow the searches to proceed until the upper bounds on iteration counts
have been reached.
For this task, we split the set of instances into 500 validation and 500 test
instances (see \cref{sec:data_splits} for details).

The results shown in \cref{tab:planning_results}
continue to demonstrate a significant performance for \methodname\ over
baseline strategies, achieving an 85.0\% Success Rate on the validation set
and 83.8\% on the test set.
Notably, the two-stage approach using Gemini 1.5 Pro achieves
success rates to 98.4\% and 98.2\% on validation and test respectively.

\begin{figure}[t]
    \centering
    \includegraphics[width=0.48\textwidth]{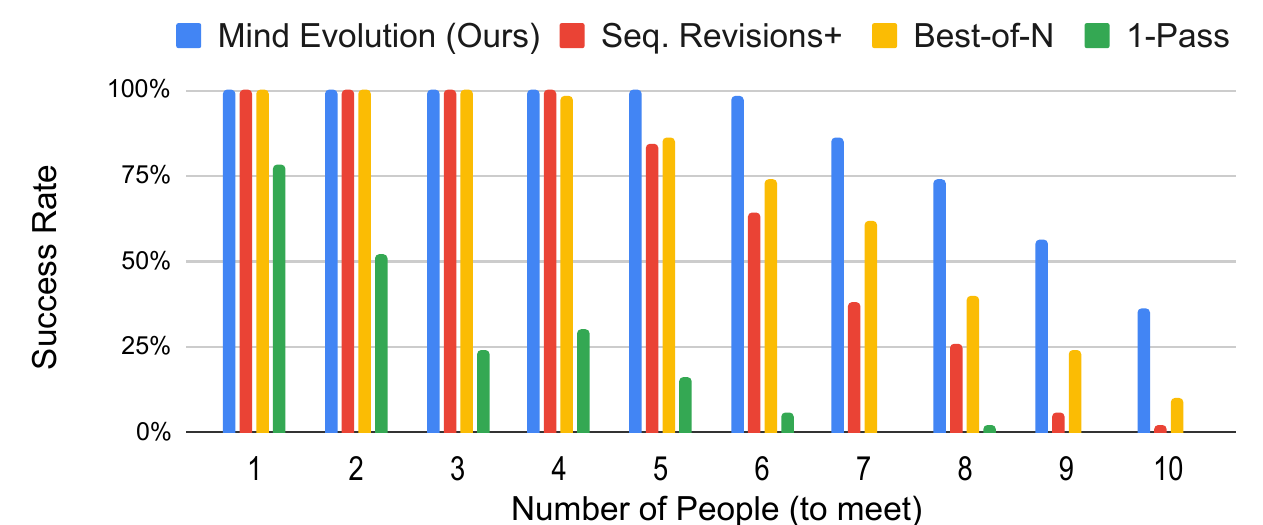}
    \caption{Success rate on the validation set of the Meeting Planning benchmark per number of people to meet with.}
    \label{fig:hist_meeting_planning}
\end{figure}

Finally, \cref{fig:hist_meeting_planning} shows the breakdown of success rates
by the number of people to schedule meetings with.
In this case, we find that \methodname\ sustains a significant advantage in
success rate as the number of people increases. 

\subsection{Analysis and Ablation Studies}
\label{subsec:ablations}

To understand how \methodname's performance scales, and how the different 
components affect its behavior,
we provide additional measurements and ablations to gain additional insight.

\begin{figure}[t]
    \centering
    \includegraphics[width=0.45\textwidth]{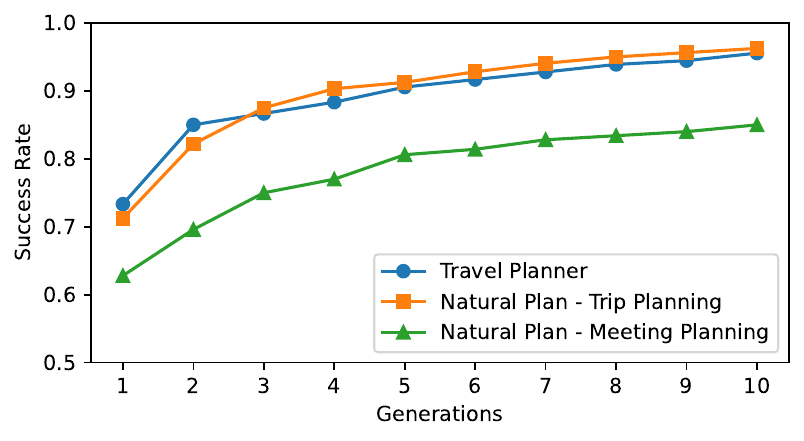}

    \caption{Success rate on the validation set for each natural language planning benchmark at each generation of \methodname.}
    \label{fig:generations_performance_planning_tasks}
\end{figure}

\paragraph{Scaling}
Regarding scaling, \Cref{fig:generations_performance_planning_tasks} 
reports the Success Rate achieved by \methodname\ across
the planning tasks as a function of the number of generations.
These results clearly show steady improvement for
\methodname\ as the number of generations is increased.

\begin{figure}[h!]
    \centering
    \includegraphics[width=0.48\textwidth]{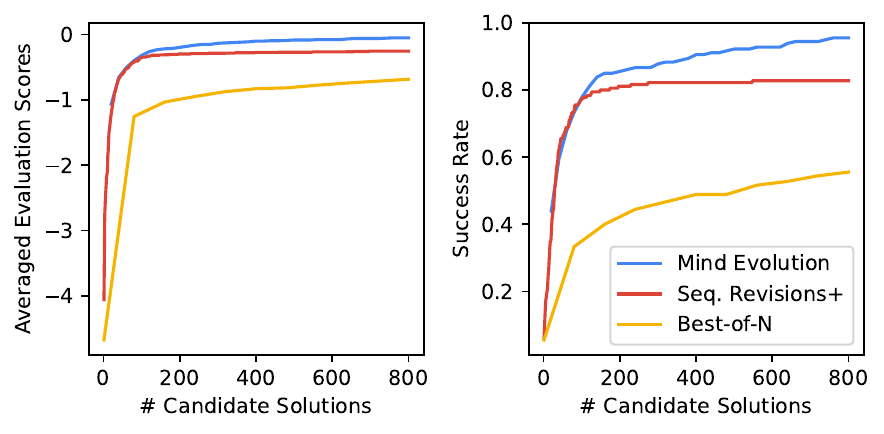}
    \caption{TravelPlanner success rates and evaluation scores as the number of candidate solutions is increased.}
    \label{fig:travel_planner_trends}
\end{figure}

\begin{figure}[h!]
    \centering
    \arrayrulecolor{black}
    \includegraphics[width=0.48\textwidth]{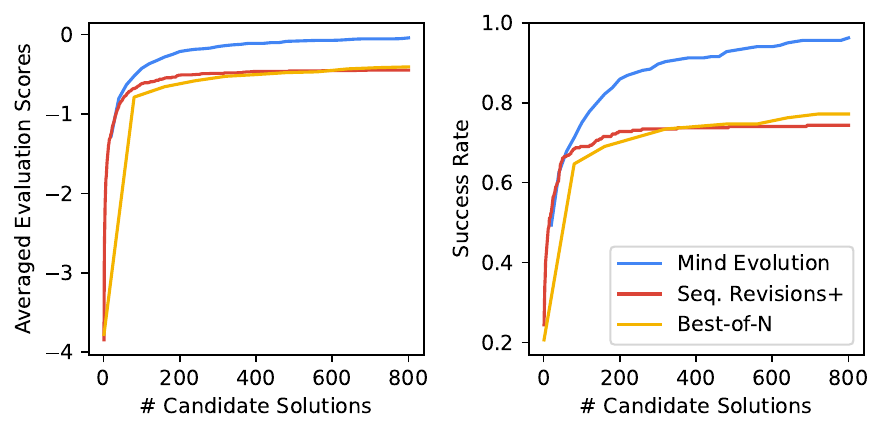}
    \caption{Trip Planning success rates and evaluation scores as the number of candidate solutions is increased. 
    }
    \label{fig:np_trip_planning_trends}
\end{figure}

\begin{figure}[h!]
    \centering
    \includegraphics[width=0.48\textwidth]{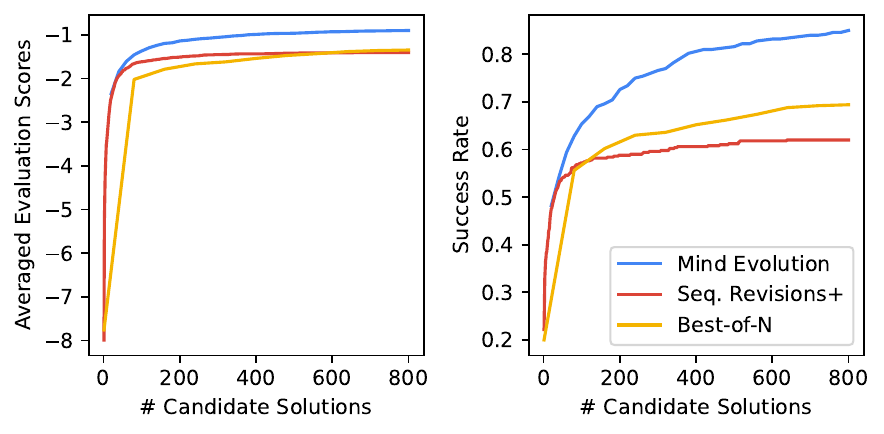}
    \caption{Meeting Planning success rates and evaluation scores as the number of candidate solutions is increased.}
    \label{fig:np_meeting_planning_trends}
\end{figure}

To compare the scaling of \methodname\ to that of the baseline search methods,
we also plot the Success Rate and average task evaluation scores
as a function of the number of candidate solutions generated
by the each strategy
(Figures \ref{fig:travel_planner_trends}--\ref{fig:np_meeting_planning_trends}).
The task evaluation scores are calculated by penalizing unsatisfied constraints
and suboptimality of the objective value,
hence the maximum score that can be achieved in any problem instance is zero
(see \cref{subsec:eval_func} for details).
In \cref{sec:additional_results_price}, we provide another perspective on the
cost-benefit trade-offs in terms of the specific API costs incurred.

Figures \ref{fig:travel_planner_trends}--\ref{fig:np_meeting_planning_trends}
show the results for the TravelPlanner, Trip Planning and Meeting Planning 
tasks respectively.
In each case,
we see that the overall success rates and average task evaluation scores 
improve monotonically with an increasing number of proposed solutions
across all search methods.
These plots also show that \methodname\ is consistently more effective than
the baseline strategies with respect to the number of candidate solutions
needed to achieve a specified level of success rate 
(or average task performance).

We note that Best-of-N appears to be significantly underperforming on 
TravelPlanner.
We hypothesize that this occurs because this task involves implicit 
commonsense constraints
(e.g., a trip plan should return to the origin city,
a restaurant cannot be visited twice, etc.),
which are not given in the problem instance but instead learned from evaluation
feedback, which Best-of-N does not leverage.

\paragraph{Ablations}
We also conducted a set of ablations to study the contribution of the
different components of \methodname.
\cref{tab:prompt_ablation} shows that
using the critic step in the \rcc\ process
(\Cref{fig:critical_thinking} in \Cref{subsec:our_method})
and textual feedback from the evaluation functions
are the most critical to performance, 
although the other components also make meaningful contributions to performance.

\begin{table}[h!]
    \setlength{\tabcolsep}{0.4em}
    \footnotesize
    \centering
    \begin{tabular}{|r|ccccc|}
        \hline
        Critic & & \checkmark & \checkmark & \checkmark & \checkmark  \\
        S/Q Prompts & & & \checkmark & \checkmark & \checkmark \\
        Textual Feedback & & & & \checkmark & \checkmark \\
        Reset with LLM & & & & & \checkmark \\
        \hline
        Success Rate & $46.1\%$ & $71.1\%$ & $76.1\%$ & $91.1\%$ & $95.6\%$ \\
        \hline
    \end{tabular}
    \caption{
An ablation study of Mind Evolution components on the TravelPlanner
validation set.
Each column in the table shows an experiment where \checkmark\ indicates
whether a component is used.
If ``Critic'' is disabled, we skip the critic step in
\cref{fig:critical_thinking} and go straight to the author step.
``S/Q Prompts'' stands for Strategy/Question prompts,
which are additional task-specific instructions in the critical thinking
prompts (see \cref{subsec:prompt_design} for details).
If ``Textual Feedback'' is disabled, we do not include evaluation feedback
in the prompts.
If ``Reset with LLM'' is disabled, we directly select global elites by their
evaluation scores in island reset events, rather than use an LLM to choose,
as described in \Cref{subsec:our_method}.}
    \label{tab:prompt_ablation}
\end{table}

\begin{table}[!ht]
    \setlength{\tabcolsep}{0.3em}
    \small
    \centering
    \begin{tabular}{|l|r|}
        \hline
         & Succ.\ Rate \\
        \hline
        w/ island model ($N_{\text{island}}=4$, $N_{\text{convs}}=5$) & 87.5\% \\
        w/o island model ($N_{\text{island}}=1$, $N_{\text{convs}}=20$) & 77.4\%\\
        \hline
        \hline
        $N_{\text{convs}}=10, N_{\text{gens}}=5$ & 82.5\% \\
        $N_{\text{convs}}=5, N_{\text{gens}}=10$ (default) & 87.5\% \\
        $N_{\text{convs}}=4, N_{\text{gens}}=13$ & 85.0\% \\
        \hline
    \end{tabular}
    \caption{
Hyperparameter studies on the Trip Planning problem instances with 10 cities.
The first two rows show the difference between enabling and disabling 
the island model.
The bottom three rows illustrate a trade-off between the number of candidates
per generation versus the number of generations.
(Note that the bottom row ($N_{\text{convs}}=4, N_{\text{gens}}=13$) produces
slightly more than 800 solutions (832).}
    \label{tab:hyperparameter_study}
\end{table}

\begin{figure*} %
\lstset{
  basicstyle=\ttfamily\scriptsize,
  breaklines=true
}

\renewcommand{\tabcolsep}{2pt}

\begin{tabular}{c||c|l}
Task&
\multicolumn{2}{c}{Results (Cipher and Poem)}\\
\hline
\begin{minipage}{0.3\textwidth}
\begin{lstlisting}
Message To Encode (M): 
            10,20,30,40,
            50,60,70,80,
            90,100,10,20
            
Style: Poem
Inspiration: Shel Silverstein
Words Between (B): 4
Topic: "The Fun of Walking"
\end{lstlisting}
\end{minipage} &
\begin{minipage}{0.16\textwidth}
\begin{lstlisting}
10 : rooster;
20 : flowers;
30 : bright;
40 : flames;
50 : cherry;
60 : crimson;
70 : sunset;
80 : ruby;
90 : scarlet;
100 : burning;
\end{lstlisting}
\end{minipage}  &
\begin{minipage}{0.50\textwidth}
\begin{lstlisting}
I like to walk, I like to stride,
With ROOSTER crows and FLOWERS by my side. 
I like to walk, I like to roam, 
Past BRIGHT green fields and FLAMES at home.
I like to walk, I like to stroll,
To see the CHERRY trees and a CRIMSON hole.
I like to walk, I like to ramble,
To watch the SUNSET and hold a RUBY's gamble.
I like to walk, I like to wander,
To watch a SCARLET leaf, and a fire, BURNING, under.
I like to walk, I like to stride,
With ROOSTER crows and FLOWERS by my side. 
\end{lstlisting}
\end{minipage} \\
\end{tabular}

\caption{%
    \textbf{\steg\ example.}
Example of the encoding of a \steg\ problem instance (left)
and a correct solution (right) that includes the number-to-word cipher and 
a poem in the style of a children's poetry author.
Note that $|M| = 12$ in this instance.
We added capitalization to the code words to highlight them.
}
\label{fig:stegExample}
\end{figure*}

\begin{table*} %
    \setlength{\tabcolsep}{0.4em}
    \footnotesize
    \centering
    \begin{tabular}{|r|c|r|r|r|r|}
        \hline
         & Set & Success Rate & Input Tokens & Output Tokens & API Cost (Oct 2024) \\
        \hline
        1-Pass & val & $0/101=0.0\%$ & $0.002$M & $<0.001$M & $<$\$$0.001$ \\
        Best-of-N & val & $1/101=1.0\%$ & $1.56$M & $0.25$M & \$$0.19$ \\
        \sequential & val & $20/101=19.8\%$ & $41.69$M & $0.24$M & \$$3.20$  \\
        \bf{\methodname} & val & $47/101=\textbf{46.5\%}$ & $3.56$M & $0.20$M & \$$0.33$ \\
        \bf{(+pro)} & val & $88/101=\textbf{87.1\%}$ & $3.74$M & $0.22$M  & \$$0.65$ \\
        \hline
        \bf{\methodname} & test & $106/245=\textbf{43.3\%}$ & \$$0.34$ & $3.63$M & $0.22$M \\
        \bf{(+pro)} & test & $194/245=\textbf{79.2\%}$ & \$$0.72$ & $3.84$M & $0.24$M \\
        \hline
    \end{tabular}
    \caption{%
        \textbf{Experimental results on \steg.}
        Price and token counts are averages per problem.
        All results use Gemini 1.5 Flash, except
        \textbf{(+pro)}, which solves the problems that were not solved in the Flash\ runs, using Gemini 1.5 Pro.
    }
    \label{tab:experiments_steg_test}
\end{table*}

\begin{figure} %
    \centering
    \includegraphics[width=0.48\textwidth]{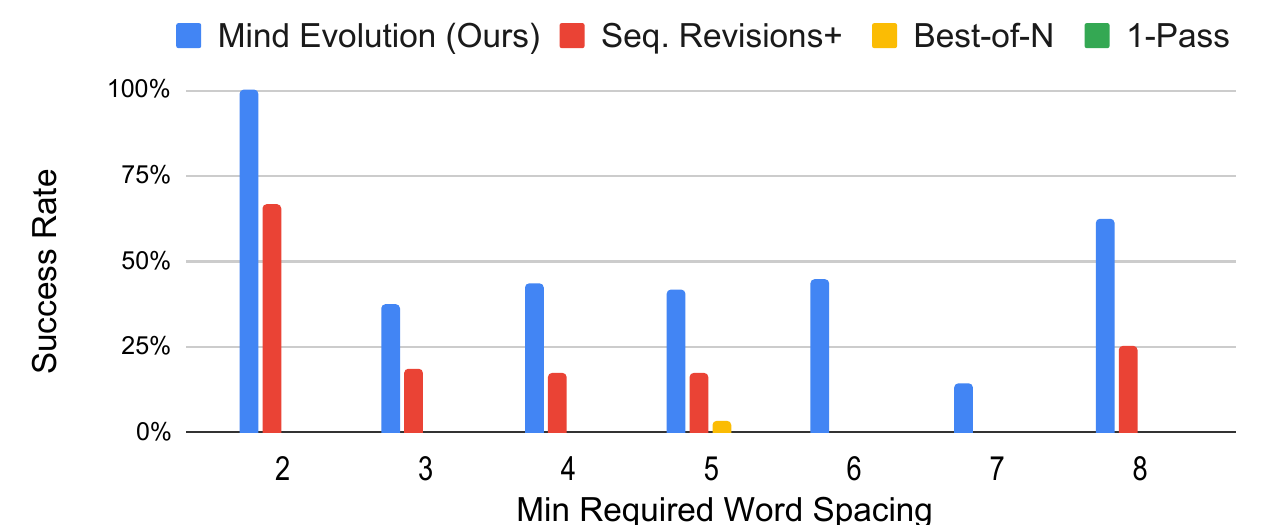}
    \caption{
Histogram of Success Rate for each difficulty level.
1-Pass returns valid responses, but fails to solve any of the problems,
so it is not visible in the histogram.
}
    \label{fig:hist_steg}
\end{figure}

To assess hyperparameter sensitivity, we investigated the Trip Planning task 
in greater detail, choosing the harder setting with 10 cities 
to better reveal differences in performance.
(Similar results are also observed on the harder problem instances
from the other benchmark tasks.)
In \cref{tab:hyperparameter_study},
the top two rows compare the effect of including or excluding the island
model from the evolutionary search, controlling for the same number
(800) of candidate solutions.
These results show that the island model significantly improves the performance
of \methodname.
The bottom three rows compare
the effect of increasing the number of candidate solutions per generation
versus having more generations 
while controlling for a similar number of candidates considered overall.
In this case, it appears that deeper evolutionary search indeed has benefits,
although it is also important to continue exploring broadly in each generation.

\ifsteg
\section{A Challenging New Task: \steg}
\label{sec:steg}

We introduce a challenging new task, \steg,
where a hidden message should be stenographically encoded \citep{provos2003hide}
into a piece of creative writing.
Even though the problem is difficult to formalize,
it remains amenable to programmatic verification, 
which makes it addressable by the methods considered in this paper.
In this task, a hidden message ($M$) expressed by a sequence of numbers
should be encoded in a piece of creative text about a particular topic,
expressed in the form of an essay, story or poem. 
The goal is to both provide a number-to-word substitution cipher
and a generated text that uses the cipher to encode the message.
\cref{fig:stegExample} gives an example.
We impose an additional constraint that there must be, on average, $B$
words between successive cipher words in the generated text,
which ensures
that simply listing the cipher words as the text portion does not qualify as
solution when $B > 0$.

The difficulty of this problem varies along four axes:
\begin{enumerate}[leftmargin=1em,nosep]
    \item Difficulty increases with the length of the hidden message, $M$.
        We set $10 \leq |M| \leq 30$. 
    \item The repetition of the numbers in $M$.
        The more repetition, the more stringent the constraints.
    \item The ``closeness'' of the repeated numbers to each other.
        Each form of writing dictates how much repetition of the same word and
proximity of occurrence is acceptable.
        The LLM must balance adherence to the form with the need to correctly
encode the message.
    \item Empirically, as $B$ (the mean distance between cipher words) grows, the problem becomes more difficult.
        In our tests, $3 \leq B \leq 7$.  
\end{enumerate}
We divide the problem instances into
a validation split of 101 instances and a test split of 245 instances.
See \cref{sec:additional_steg} for additional details about the \steg\ evaluation.

Detailed performance results for \methodname\ and the baseline strategies
are given in \cref{tab:experiments_steg_test},
while \cref{fig:hist_steg} shows performance per difficulty level.
Here the two-stage \methodname\ (+pro) achieves 87.1\% on validation and 79.2\% on test.
Best-of-N only manages to solve 1\% of the validation tasks.

\fi

\section{Conclusion}
\label{sec:conclusion}

We have presented \methodname, an evolutionary search approach for solving
challenging natural language planning problems,
by scaling inference-time compute for stochastic exploration and iterative
refinement.
An evaluation on the TravelPlanner and Natural Plan natural language planning
benchmarks%
\ifsteg
, as well as a new benchmark \steg\ introduced in this paper,
\fi
demonstrates that \methodname\ significantly outperforms Best-of-N
and sequential revision.
To our knowledge, this is the first approach that is able to achieve such a
level of success on these tasks without explicitly leveraging a formal solver.

\paragraph{Limitations}
The main limitation of the current work is the focus on natural language
planning problems where proposed solutions can be programmatically 
evaluated and critiqued.
In future work, we aim to extend beyond this limitation by developing
LLM-based evaluators that would enable broader applications.

\section*{Acknowledgement}
The authors thank Sergio Guadarrama and Doina Precup for supporting this work.
We also thank Sirui Xie, John Canny, and the Google DeepMind FunSearch team for valuable discussion.

\bibliography{main}

\clearpage
\appendix

\section{Implementation Details}
\label{sec:implementation_details}

Here we describe the implementation details of \methodname.
The code will be made available.

\subsection{Prompt Design}
\label{subsec:prompt_design}

\begin{figure*}[t]
    \centering
    \includegraphics[width=0.95\textwidth]{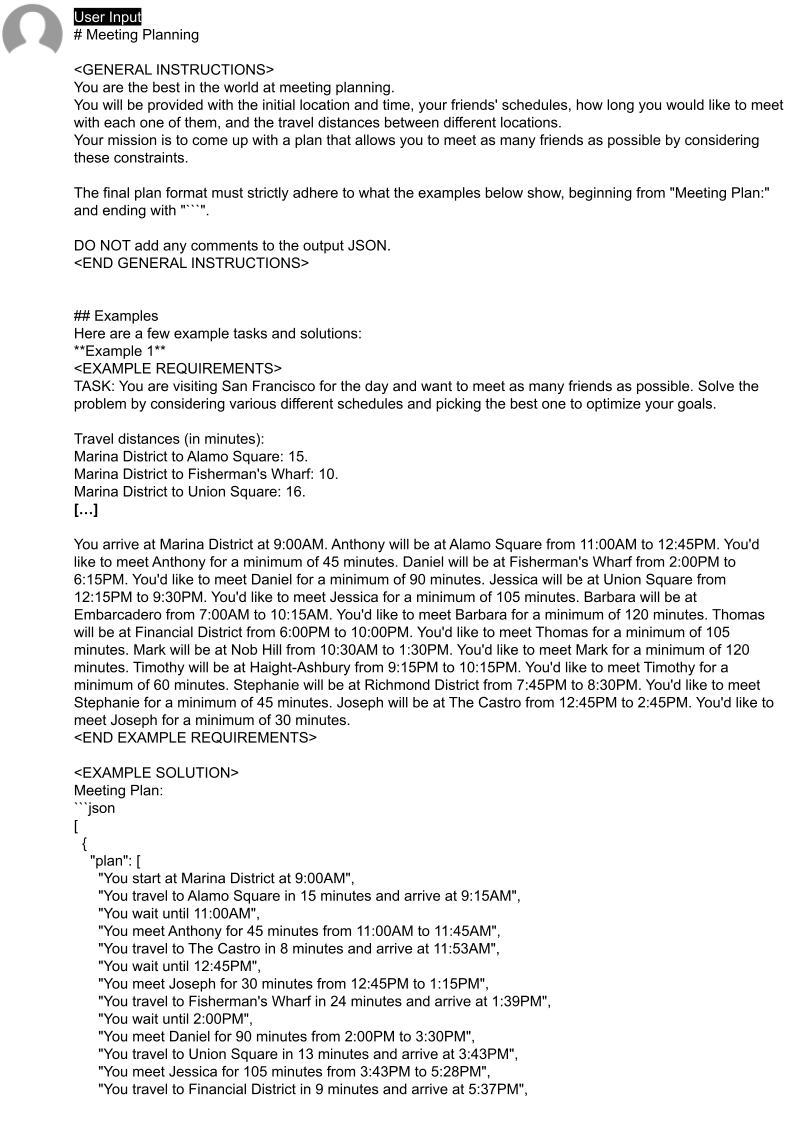}
    \caption{Example Meeting Planning prompt and model response with parent solutions given (Part 1)}
    \label{fig:ex_meeting_planning_prompt_1}
\end{figure*}

\begin{figure*}[t]
    \centering
    \includegraphics[width=0.95\textwidth]{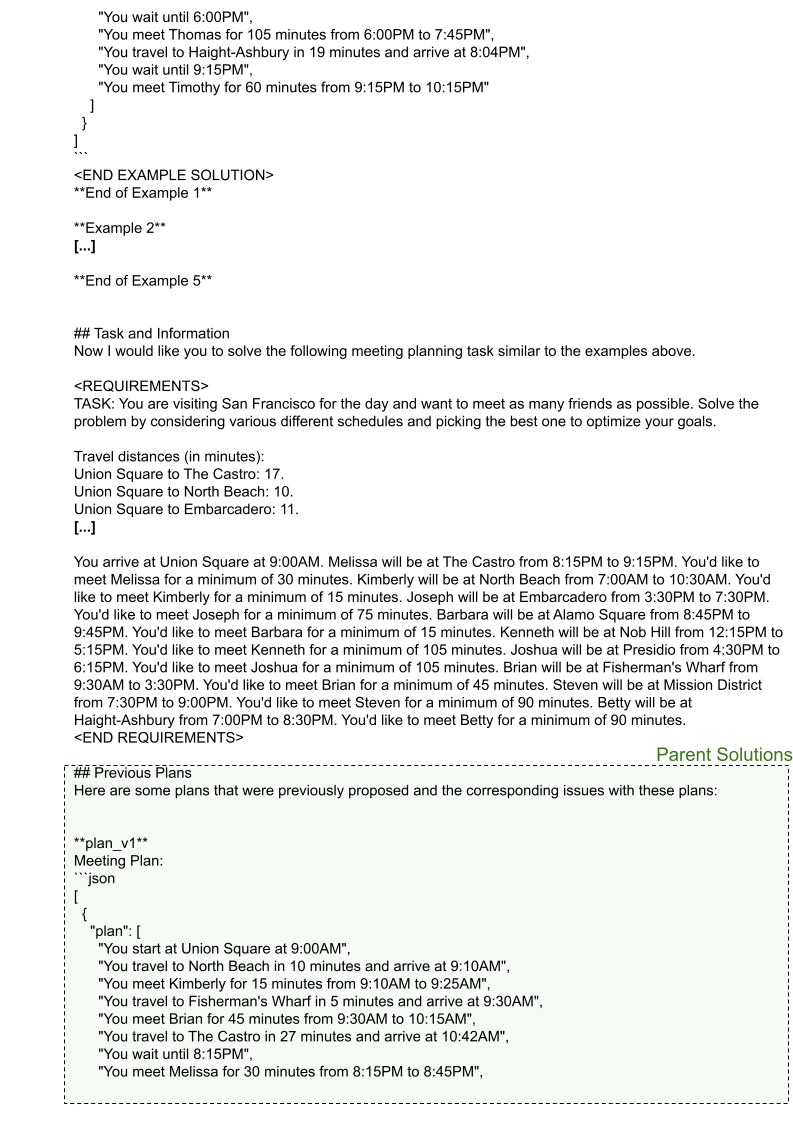}
    \caption{Example Meeting Planning prompt and model response with parent solutions given (Part 2)}
    \label{fig:ex_meeting_planning_prompt_2}
\end{figure*}

\begin{figure*}[t]
    \centering
    \includegraphics[width=0.95\textwidth]{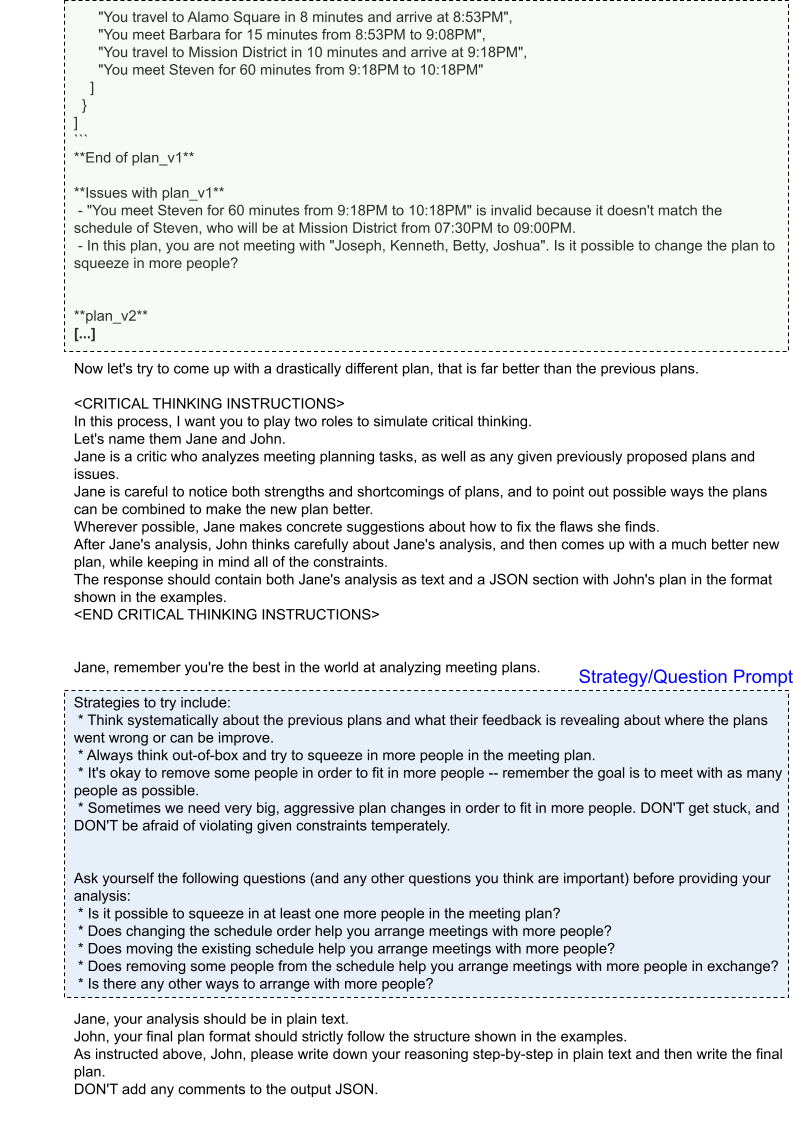}
    \caption{Example Meeting Planning prompt and model response with parent solutions given (Part 3)}
    \label{fig:ex_meeting_planning_prompt_3}
\end{figure*}

\begin{figure*}[t]
    \centering
    \includegraphics[width=0.95\textwidth]{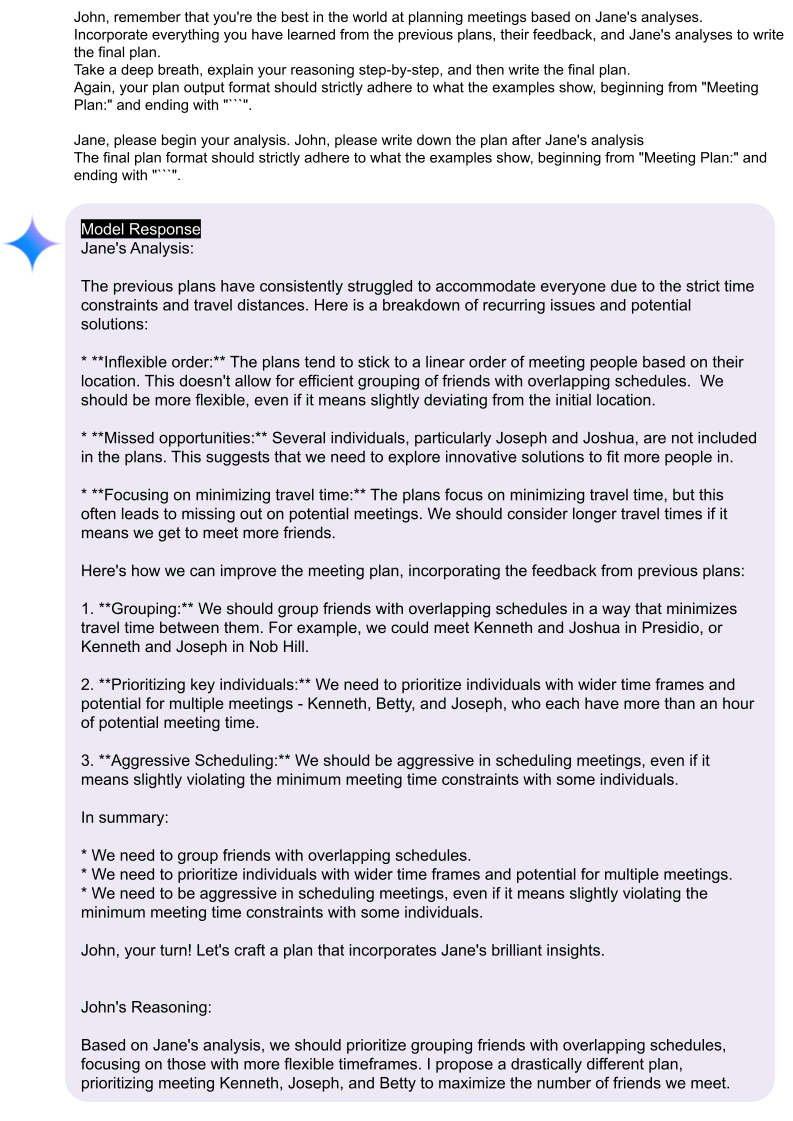}
    \caption{Example Meeting Planning prompt and model response with parent solutions given (Part 4)}
    \label{fig:ex_meeting_planning_prompt_4}
\end{figure*}

\begin{figure*}[t]
    \centering
    \includegraphics[width=0.95\textwidth]{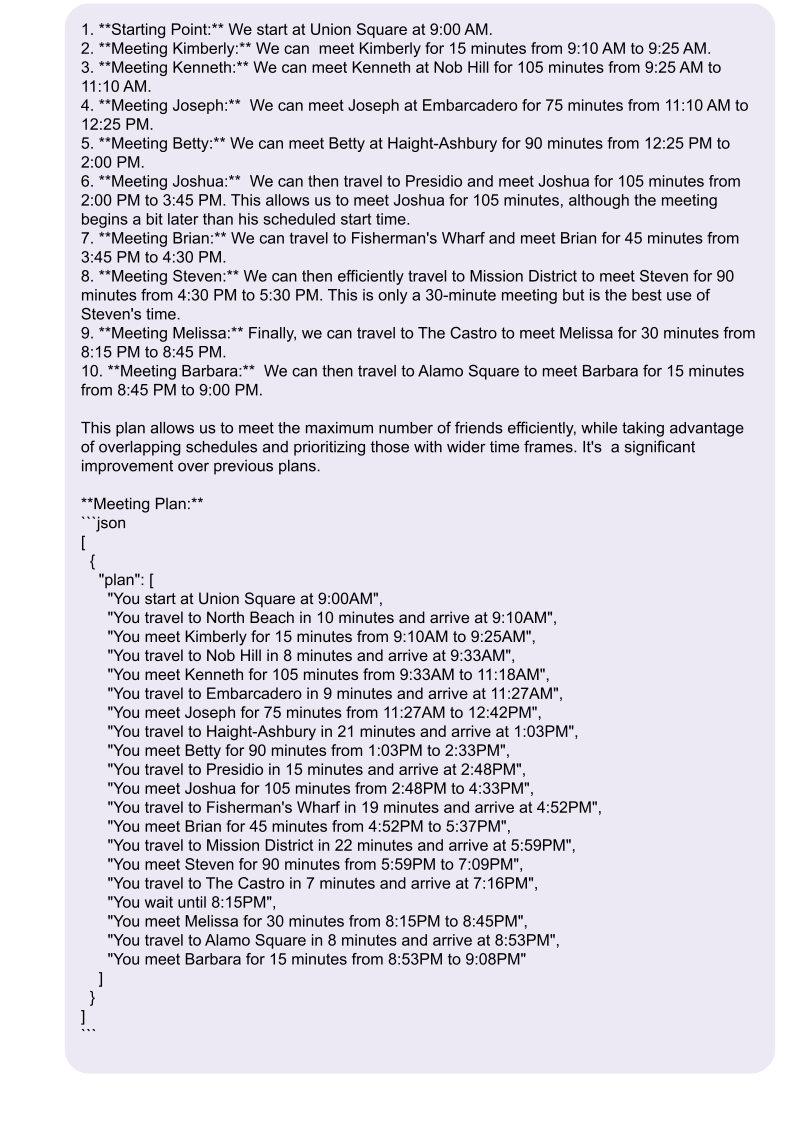}
    \caption{Example Meeting Planning prompt and model response with parent solutions given (Part 5)}
    \label{fig:ex_meeting_planning_prompt_5}
\end{figure*}

We first use Meeting Planning as an example to illustrate the structure of the prompts used.
The prompts, as well as the model responses when parent solutions are given, are shown in Figures \ref{fig:ex_meeting_planning_prompt_1}-\ref{fig:ex_meeting_planning_prompt_5}.
The prompts begin with general instructions
and a general problem definition,
few-shot examples, then a task description.
The few-shot examples help the LLM understand the problem and generate
solutions closer to the desired formats.
For TravelPlanner, we take two 3-day example plans from the training set and
use them across all tasks (3-7 days).
For Trip Planning, we take two example plans from the few-shot examples
provided by the benchmark and use them across all tasks.
For Meeting Planning, we use the 5-shot examples provided by the benchmark
for each task.

After the task description,
we include parent solutions with corresponding evaluation feedback,
followed by critical thinking instructions
(in Figures \ref{fig:ex_meeting_planning_prompt_3}--\ref{fig:ex_meeting_planning_prompt_4}).
These instructions lead the LLM to improve the parent solutions,
following the \refine\ (\rcc) process described in \cref{subsec:our_method}.
The critical thinking instructions include problem-specific 
Strategy/Question prompts based on findings in each
validation set (ablated in \cref{subsec:ablations}).
In the model responses, one can see that the LLM follows the critical
thinking instructions in playing the critic role to analyze the parent
solutions, and playing the author role to propose a new solution. 

\begin{figure*}[t]
    \centering
    \includegraphics[width=0.95\textwidth]{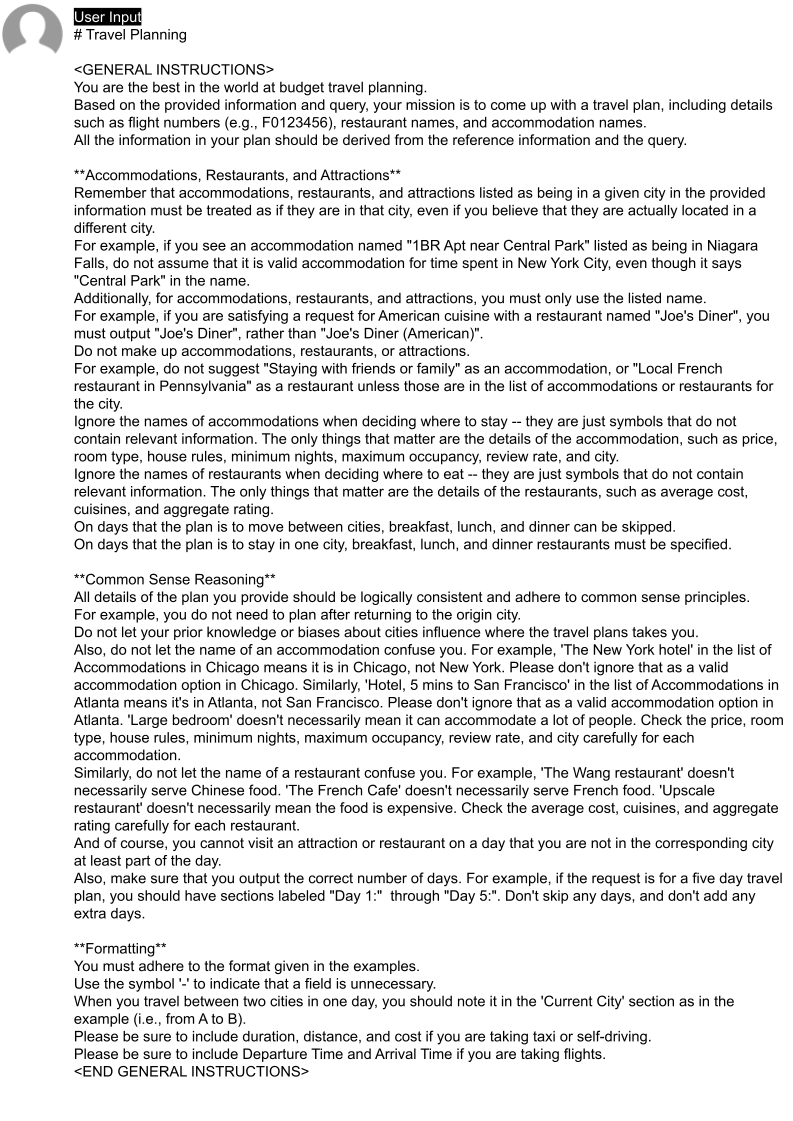}
    \caption{Example TravelPlanner prompt and model response with parent solutions given (Part 1)}
    \label{fig:ex_travelplanner_prompt_1}
\end{figure*}

\begin{figure*}[t]
    \centering
    \includegraphics[width=0.95\textwidth]{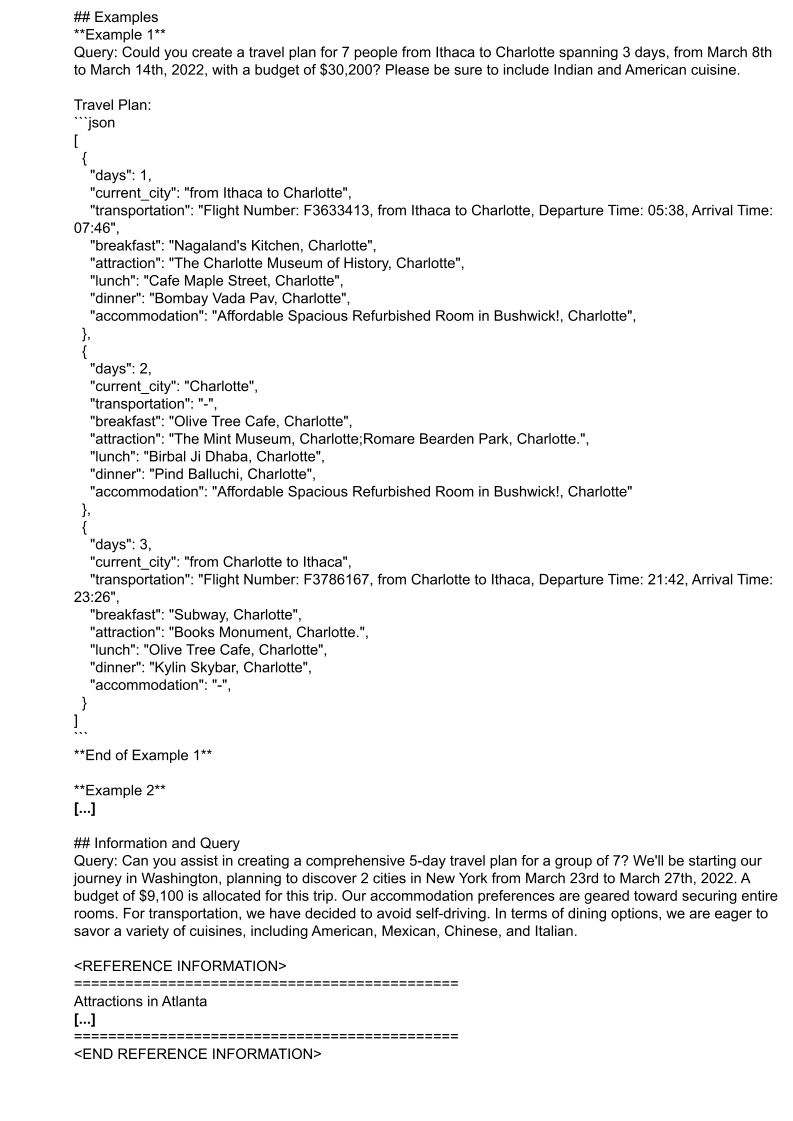}
    \caption{Example TravelPlanner prompt and model response with parent solutions given (Part 2)}
    \label{fig:ex_travelplanner_prompt_2}
\end{figure*}

\begin{figure*}[t]
    \centering
    \includegraphics[width=0.95\textwidth]{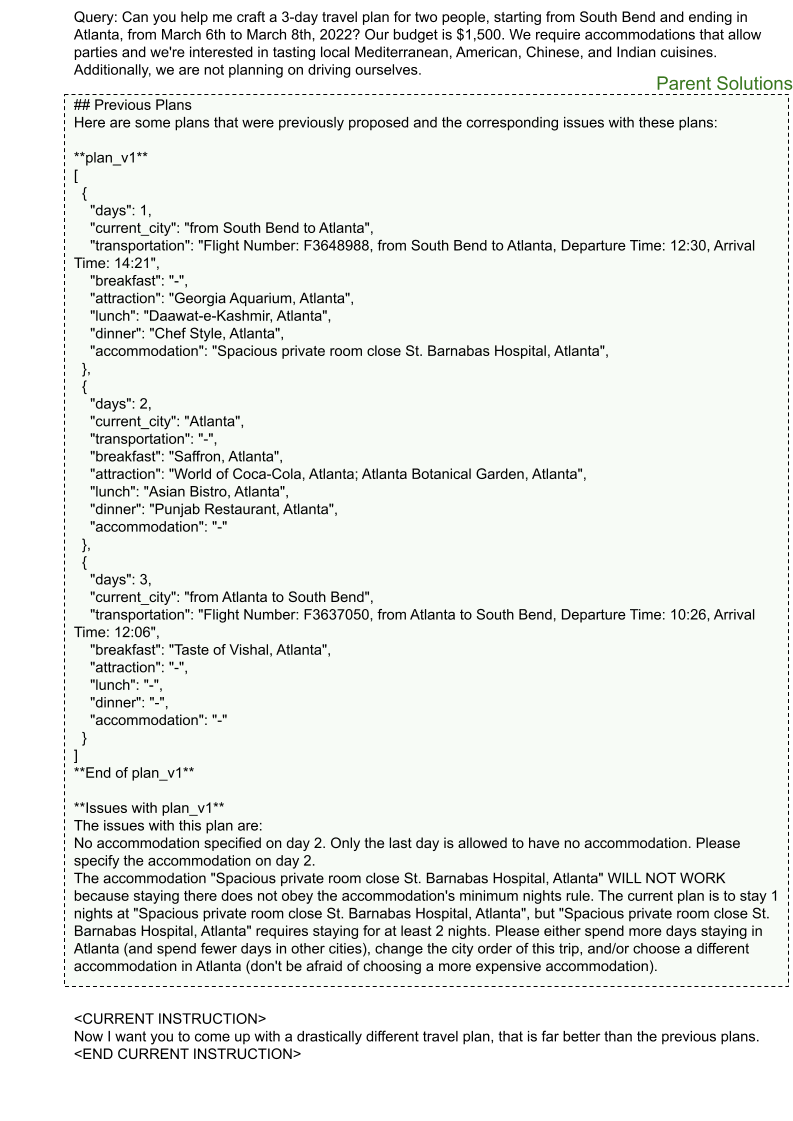}
    \caption{Example TravelPlanner prompt and model response with parent solutions given (Part 3)}
    \label{fig:ex_travelplanner_prompt_3}
\end{figure*}

\begin{figure*}[t]
    \centering
    \includegraphics[width=0.95\textwidth]{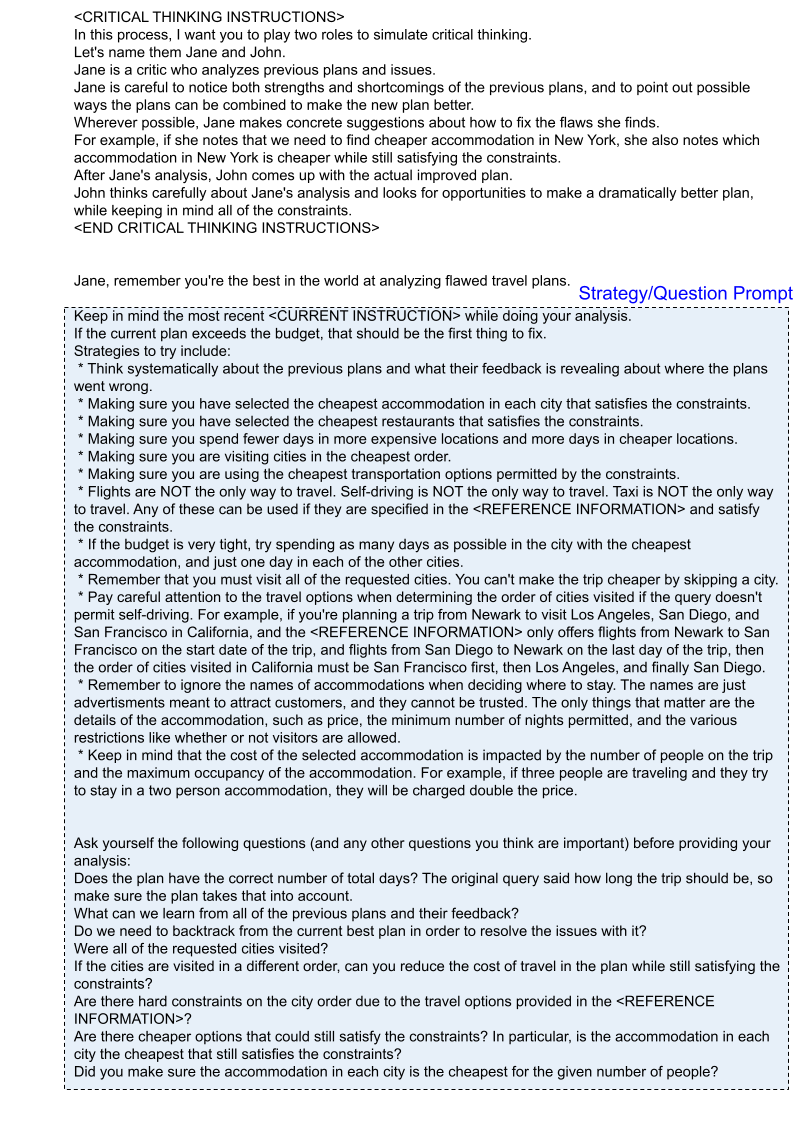}
    \caption{Example TravelPlanner prompt and model response with parent solutions given (Part 4)}
    \label{fig:ex_travelplanner_prompt_4}
\end{figure*}

\begin{figure*}[t]
    \centering
    \includegraphics[width=0.95\textwidth]{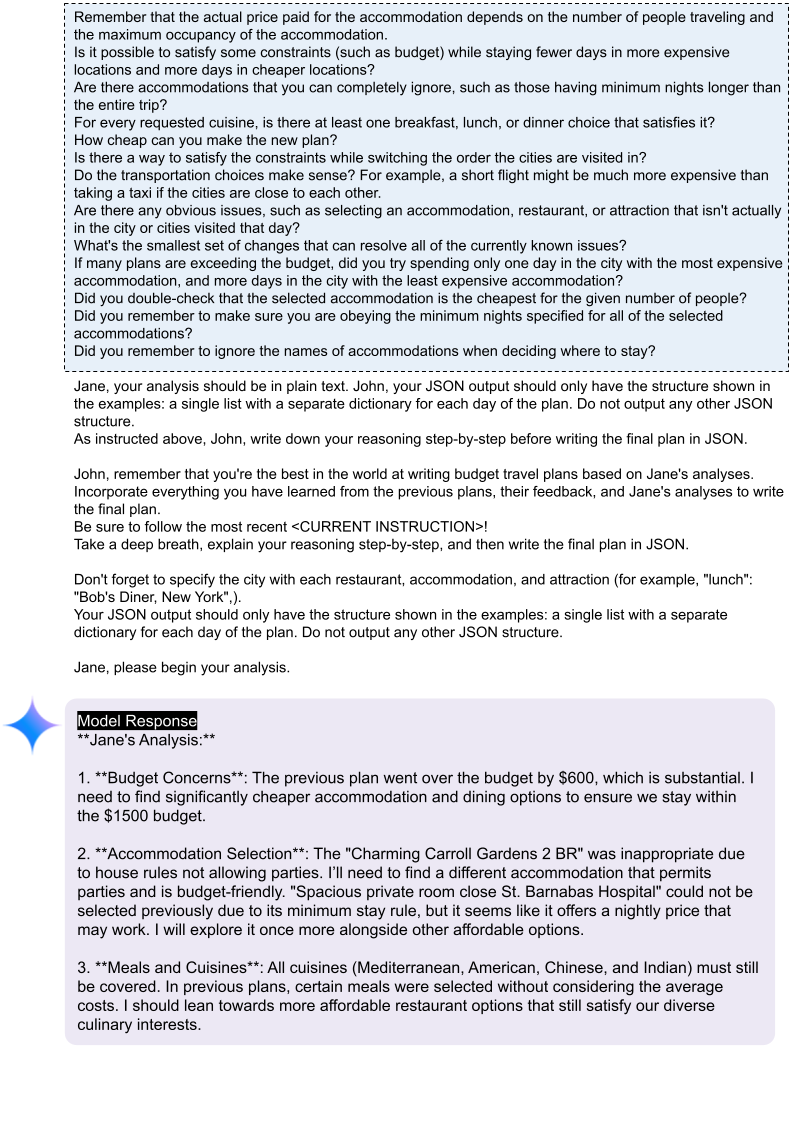}
    \caption{Example TravelPlanner prompt and model response with parent solutions given (Part 5)}
    \label{fig:ex_travelplanner_prompt_5}
\end{figure*}

\begin{figure*}[t]
    \centering
    \includegraphics[width=0.95\textwidth]{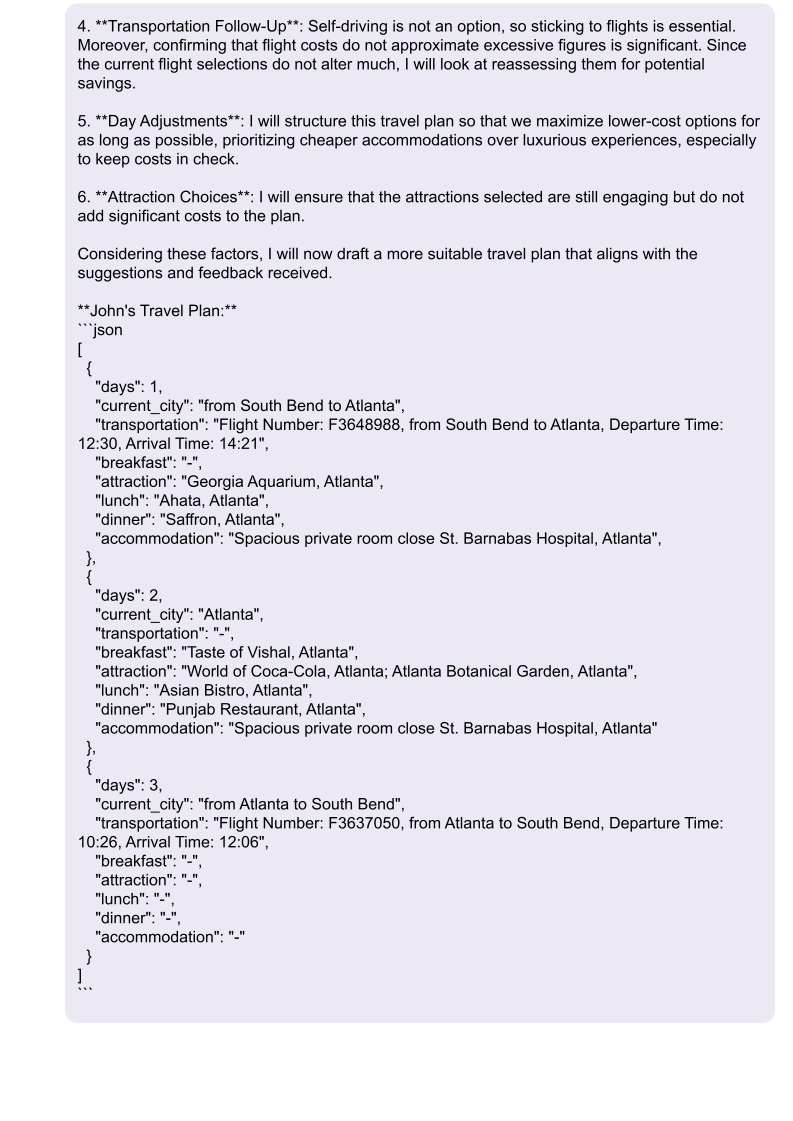}
    \caption{Example TravelPlanner prompt and model response with parent solutions given (Part 6)}
    \label{fig:ex_travelplanner_prompt_6}
\end{figure*}

We also give an example of the prompt and a model response
for TravelPlanner,
which has the same structure,
in Figures \ref{fig:ex_travelplanner_prompt_1}--\ref{fig:ex_travelplanner_prompt_6}.

\subsection{Evaluation Functions}
\label{subsec:eval_func}

In this work, solutions are evaluated programmatically with a function.
As described in \cref{subsec:our_method},
an evaluation function has three main roles:
(1) scoring solutions by measuring the optimization objective, if any;
(2) verifying whether the solution satisfies given constraints; 
and (3) providing corresponding textual feedback.
Specifically, we score natural language plans by penalizing the constraints
that are not satisfied, the objectives that are not maximized,
and for not following the required solution format.
Thus the maximum score for all tasks is zero.
We also provide textual feedback that describes how the constraints are not
satisfied and how the objectives are not maximized.

\paragraph{TravelPlanner}
Our evaluation function for TravelPlanner is modified from the TravelPlanner
evaluation code~\cite{xie2024travelplanner}.
The evaluation code expects travel plans in JSON format.
We modify the original evaluation code to make it output a
cumulative score that reflects all the constraints that are not satisfied,
instead of simply answering whether or not a plan satisfies all the constraints.
We also make it provide textual feedback for the violated constraints.

In the TravelPlanner validation set, the constraints are provided in both user
query text and a structured JSON format.
However, in the test set, the constraints are only described in user query text.
To make it easier for the evaluation function to consider the constraints,
we extract them from user query into JSON using Gemini 1.5 Flash.
For example, to extract the requested cuisines, we prompt Gemini with 
``Look at the following text and tell me if there are any cuisine requirements
on the upcoming trip...''
multiple times, and formulate the final answer via majority voting.
To verify the reliability of this approach, we tested on the validation set
and found complete agreement between the JSON extracted from user query and
the provided JSON.
In addition, we upload our test solutions to the TravelPlanner evaluation
server, and found that the results agree with the official evaluation.

\paragraph{Trip Planning} 
Similar to TravelPlanner, the Trip Planning evaluation function expects plans
in JSON format.
Since Trip Planning user queries are programmatically generated,
we can parse the constraints specified in user queries.
These constraints include number of days to stay in a city,
specific days to be in a city (e.g., for events), and whether there are
flights between cities.
Our evaluation function scores a plan by the constraints that are not satisfied
and whether it conforms with the desired JSON format,
while also providing corresponding textual feedback.

\paragraph{Meeting Planning}
The Meeting Planning evaluation function also expects plans in JSON.
Constraints are also provided in structured JSON format.
Unlike TravelPlanner and Trip Planning, Meeting Planning has an optimization
objective -- the number of friends to meet with.
We modify the original evaluation evaluation function to score a proposed plan
by how many people that are not going to be met with;
whether it conflicts with the schedules of other people;
whether it includes meetings with the same person more than once;
whether any part of the plan conflict with other parts;
whether it follows the desired format as instructed.
In Figures \ref{fig:meeting_planning_eval_func_1}--\ref{fig:meeting_planning_eval_func_2}
we present the evaluation function that implements the simple logic described
above as an example.

\clearpage

\clearpage

\begin{figure*}[t]

\noindent\begin{tcolorbox}[
    colframe=darkgray, %
    boxrule=0.3pt, %
    colback=green!5, %
    arc=3pt, %
    fontupper=\small,
    ]

\begin{lstlisting}[
    breaklines=true,
    basicstyle=\footnotesize
]
import datetime
from typing import Any, Sequence


def meeting_plan_eval(plan: list[str],
                      start_location: str, 
                      initial_time: str, 
                      friend_schedules: dict[str, Any], 
                      distance_matrix: dict[str, Any]):
  """Evaluate meeting plan.

  Args:
    plan: a list of planned steps, such as ['You start at Russian Hill at 9:00AM.', 'You travel to Marina District in 7 minutes and arrive at 9:07AM.', 'You wait until 3:45PM.', 'You meet James for 75 minutes from 3:45PM to 5:00PM.']
    start_location: Your initial location
    initial_time: the initial time, such as 10:30AM
    friend_schedules: friend's location, available time, the amount of time for the meeting, such as {'Stephanie': {'location': 'Mission District', 'start_time': '10:30AM', 'end_time': '1:30PM', 'meeting_time': 120}}
    distance_matrix: Distances between locations, such as {'Marina District': {'Mission District': 20}, 'Mission District': {'Marina District': 19}}
  """

  met_with = {}
  score = 0.0
  feedback = []
  cur_location = start_location
  cur_time = datetime.datetime.strptime(initial_time, "%
  assert isinstance(plan, list)

  for step in plan:
    try:
      if step.startswith("You start"):
        continue
      elif step.startswith("You travel"):
        destination = step.split("travel to ")[1].split(" in")[0].strip()
        cur_time = cur_time + datetime.timedelta(
            minutes=distance_matrix[cur_location][destination]
        )

        cur_location = destination
      elif step.startswith("You wait"):

        raw_end_time = step.split("wait until ")[1].split(".")[0].strip()
        end_time = None
        try:
          end_time = datetime.datetime.strptime(raw_end_time, "%
        except ValueError:
          score -= 2
          feedback.append(f"\"{step}\" is invalid because the time format doesn't follow the examples.")

        if end_time <= cur_time:
          end_time_str = end_time.strftime("%
          score -= 2
          feedback.append(f"\"{step}\" is invalid because but the previous step already ends at {end_time_str} and you cannot go backwards in time.")

        cur_time = end_time
      elif step.startswith("You meet"):
\end{lstlisting}
\end{tcolorbox}%
\caption{The Meeting Planning evaluation function (part 1).}
\label{fig:meeting_planning_eval_func_1}
\end{figure*}

\begin{figure*}

\noindent\begin{tcolorbox}[
    colframe=darkgray, %
    boxrule=0.3pt, %
    colback=green!5, %
    arc=3pt, %
    fontupper=\small,
    ]

\begin{lstlisting}[
    breaklines=true,
    basicstyle=\footnotesize
]
        person = step.split("meet ")[1].split(" for")[0].strip()
        if person in met_with:
          score -= 2
          feedback.append(f"\"{step}\" is invalid because you would be meeting with {person} more than once.")
        met_with[person] = 1
        new_time = cur_time + datetime.timedelta(
            minutes=friend_schedules[person]["meeting_time"]
        )

        loc = friend_schedules[person]["location"]
        start_time = friend_schedules[person]["start_time"]
        end_time = friend_schedules[person]["end_time"]
        start_time_str = start_time.strftime("%
        end_time_str = end_time.strftime("%

        if cur_location == loc and cur_time >= start_time and new_time <= end_time:
          score += 1
          cur_time = new_time
        else:
          score -= 2
          feedback.append(f"\"{step}\" is invalid because it doesn't match the schedule of {person}, who will be at {loc} from {start_time_str} to {end_time_str}.")
      else:
        raise ValueError("Unknown plan format")
    except Exception:
      score -= 10
      feedback.append(f"\"{step}\" is invalid because the format doesn't follow the examples.")

  all_names = set(friend_schedules.keys())
  not_met_with = ", ".join(list(all_names - set(met_with.keys())))

  return score, feedback

\end{lstlisting}
\end{tcolorbox}%
\caption{The Meeting Planning evaluation function (part 2).}
\label{fig:meeting_planning_eval_func_2}
\end{figure*}

\clearpage

\section{Data Splits}
\label{sec:data_splits}

\paragraph{TravelPlanner}
TravelPlanner has 45 training tasks, 180 validation tasks,
and 1,000 test tasks in the original benchmark.

\paragraph{Natural Plan -- Trip Planning}
The Trip Planning benchmakr has 1,600 example tasks.
There are eight different difficulty levels, ranging from 3 to 10 cities.
Each difficulty level has 200 examples.
We split the dataset into validation and test sets by putting the first
40 examples from each difficulty level into validation, and the last
160 examples into test, giving 320 examples in validation
(which we used for prompt development) and 1,280 for test.
In \cref{fig:hist_trip_planning},
we show the performance at each difficulty level.

\paragraph{Natural Plan -- Meeting Planning}
The Meeting Planning benchmark has 1,000 example tasks.
There are ten different difficulty levels, ranging from meeting one to ten 
different friends.
Each difficulty level has 100 examples.
We split the dataset into validation and test sets by putting the first
50 examples from each difficulty level into validation,
and the last 50 examples into test, giving 500 examples in validation
(which we used for prompt development) and 500 for test.
In \cref{fig:hist_meeting_planning},
we show the performance at difficulty level.

\section{GPT Results}
\label{sec:gpt_results}

Table~\ref{tab:gpt_results} presents the results of \methodname\ using
GPT-4o-mini with the same sets of prompts.
Specifically, with 1-pass inference, GPT-4o-mini also struggles at planning
tasks, achieving 0\% on TravelPlanner, 9.1\% success rate on Trip Planning,
and 20.2\% success rate on Meeting Planning.
Again, \methodname\ significantly improves the performance by over $100\%$
relatively across different benchmarks.

\begin{table}[h]
    \footnotesize
    \centering
    \begin{tabular}{|r|r|}
        \hline
         & Success Rate \\
        \hline
\textbf{TravelPlanner~\cite{xie2024travelplanner}} & $79.4\%$ \\
\textbf{Natural Plan~\cite{zheng2024natural} Trip Planning} & $48.1\%$ \\
\textbf{Natural Plan~\cite{zheng2024natural} Meeting Planning} & $86.4\%$ \\
        \hline
    \end{tabular}
    \caption{%
        \methodname\ with GPT-4o-Mini results on validation sets.
    }
    \label{tab:gpt_results}
\end{table}

\section{Model Pricing and API Cost Curves}
\label{sec:pricing}
\label{sec:additional_results_price}

\cref{tab:pricing} shows the API pricing of different models used in our
evaluation (Tables~\ref{tab:planning_results}), at the time of writing
(October 2024).

\begin{table}[htb]
    \small
    \centering
    \begin{tabular}{r|c|c}
        Model & Input Token & Output Token \\
        \hline
        Gemini 1.5 Flash & \$$0.075$/M & \$$0.30$/M \\
        Gemini 1.5 Pro & \$$1.25$/M & \$$5.00$/M \\
        \hline
        GPT-4o-Mini & \$$0.15$ & \$$0.60$ \\
        OpenAI o1-preview & \$$15.00$/M & \$$60.00$/M \\
    \end{tabular}
    \caption{%
Pricing at the time of writing (October 2024).
These differences serve as a proxy for real computational cost differences
among models.
    }
    \label{tab:pricing}
\end{table}

\begin{figure*}[htb]
    \centering
    \begin{subfigure}[t]{0.3\textwidth}
        \centering
        \includegraphics[width=\textwidth]{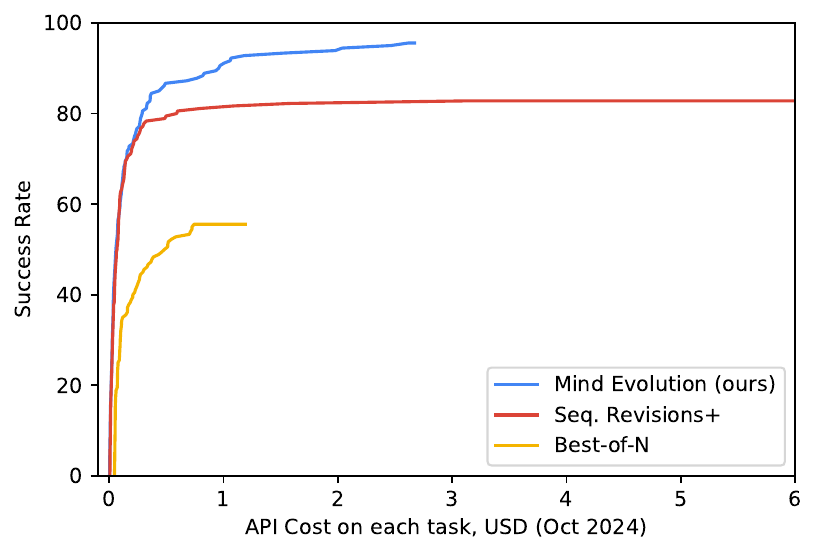}
        \caption{\textbf{TravelPlanner}}
    \end{subfigure}%
    ~ 
    \begin{subfigure}[t]{0.3\textwidth}
        \centering
        \includegraphics[width=\textwidth]{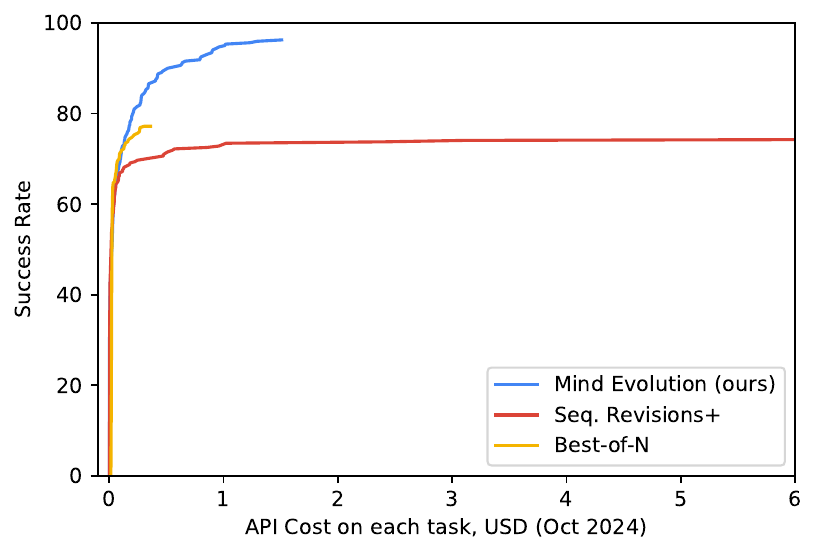}
        \caption{\textbf{Trip Planning}}
    \end{subfigure}%
    ~
    \begin{subfigure}[t]{0.3\textwidth}
        \centering
        \includegraphics[width=\textwidth]{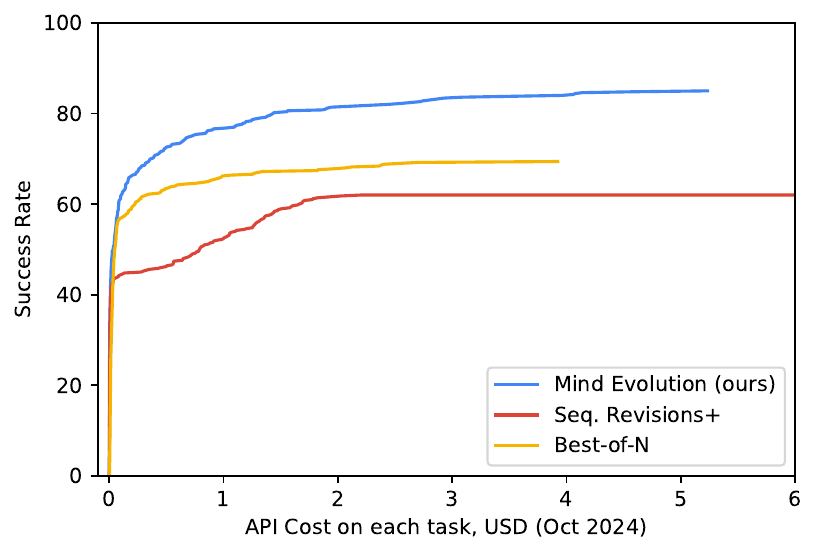}
        \caption{\textbf{Meeting Planning}}
    \end{subfigure}%
    \caption{%
        \textbf{API cost per task (Oct 2024) vs. Success Rate} on the validation sets with Gemini 1.5 Flash. API cost is also a linear combination of input token counts and output token counts. Note that \sequential\ curves are cut short. As the conversations can run up to 80 turns, it usually consumes a lot more tokens than the other methods.
    }
    \label{fig:price_performance}
\end{figure*}

\cref{fig:price_performance} gives insight into the scaling properties
of the various strategies in terms of their API cost, which is also a linear combination of the input token counts and the output token counts, weighted by base rate (\cref{tab:pricing}).

\section{Additional Examples}
\label{sec:additional_examples}

In addition to \cref{tab:trip_planning_example},
we present qualitative examples of TravelPlanner and Meeting Planning in
\cref{tab:travel_planner_example} and \cref{tab:meeting_planning_example},
respectively.

\begin{table*}[t]
    \setlength{\tabcolsep}{0.4em}
    \small
    \centering
    \arrayrulecolor{gray!70}
    \begin{tabular}{|lp{0.75\textwidth}|}
    \hline
    \multicolumn{2}{|p{0.95\textwidth}|}{
{\fontfamily{cmss}\selectfont
Q: Could you construct a 3-day journey for two people from Chicago to Albany
that takes place from March 22nd to March 24th, 2022?
Our budget is \$2,300. We require accommodations that allow smoking and should
ideally be entire rooms.
We will not be self-driving during this trip. On the subject of cuisine,
we're open to any suggestions you might have.

\vspace{+0.3em}
[Transportation Options] Taxi: N/A, Self-Driving: N/A, Flights: F3732604
\textbf{[...]}

[Accommodation Options in Albany] 2Br Union square Newly furnished
(minimum 30 nights), \textbf{[...]}

\textbf{[...]}
}} \\
& \\
\textbf{Method} & \textbf{Answer} \\
1-Pass &
{\fontfamily{cmss}\selectfont
[Day 1] city: 'from Chicago to Albany', accommodation: 'Spacious room with huge bay window \& natural light, Albany', attraction: '-', breakfast: '-', dinner: 'Kaiser, Albany', lunch: '-', transportation': 'Flight Number: F3983351, from Chicago to Albany, Departure Time: 20:20, Arrival Time: 23:04' 
[Day 2] city: 'Albany', accommodation: 'Spacious room with huge bay window \& natural light, Albany', attraction: 'Albany Pine Bush Preserve, Albany;New York State Museum, Albany;', breakfast: 'Desi Villa, Albany', lunch: 'Just Parkash, Albany', dinner: 'Kaiser, Albany', 'transportation': '-'
[Day 3] city: 'from Albany to Chicago', accommodation: '-', attraction: 'Hear about Here, Albany;', breakfast: 'Cafe Coffee Day The Square, Albany', lunch: 'PiccoLicko, Albany', dinner: 'Starvin\' Marvin, Albany', transportation: 'Flight Number: F3813300, from Albany to Chicago, Departure Time: 16:40, Arrival Time: 17:52'
\textcolor{red}{\faClose The total cost exceeds the budget of \$2300 by \$114, [...]}
} \\
Best-of-N &
{\fontfamily{cmss}\selectfont
[Day 1] city: 'from Chicago to Albany', accommodation: 'Mid Town East Side Apartment, Albany', attraction: 'New York State Museum, Albany', breakfast: '-', dinner: 'Kaiser, Albany', lunch: '-', transportation': 'Taxi, from Chicago to Albany, duration: 12 hours 53 mins, distance: 723 km, cost: 532' 
[Day 2] city: 'Albany', accommodation: 'Mid Town East Side Apartment, Albany', attraction: 'Albany Institute of History \& Art, Albany', breakfast: 'Echoes Satyaniketan, Albany', lunch: 'Cafe Coffee Day The Square, Albany', dinner: 'Urban Punjab, Albany', 'transportation': '-'
[Day 3] city: 'from Albany to Chicago', accommodation: '-', attraction: 'Historic Cherry Hill, Albany', breakfast: 'Just Parkash, Albany', lunch: 'BK\'s The Juice Bar, Albany', dinner: '-', transportation: 'Flight Number: F4008387, from Albany to Chicago, Departure Time: 05:05, Arrival Time: 06:11'
\textcolor{red}{\faClose Taxi from Chicago to Albany on day 1 is not a valid option, [...]}
} \\
Sequential Revisions+ &
{\fontfamily{cmss}\selectfont
[Day 1] city: 'from Chicago to Albany', accommodation: '2Br Union square Newly furnished', Albany', attraction: '-', breakfast: '-', dinner: '-', lunch: '-', transportation': 'Flight Number: F3983350, from Chicago to Albany, Departure Time: 14:29, Arrival Time: 17:03' 
[Day 2] city: 'Albany', accommodation: '2Br Union square Newly furnished', attraction: 'New York State Museum, Albany;Albany Pine Bush Preserve, Albany', breakfast: 'Cafe Coffee Day, Albany', lunch: 'PiccoLicko, Albany', dinner: 'Kaiser, Albany', 'transportation': '-'
[Day 3] city: 'from Albany to Chicago', accommodation: '-', attraction: 'The McPherson Legacy to the City of Albany - Robert Burns Statue, Albany', breakfast: '-', lunch: '-', dinner: '-', transportation: 'Flight Number: F3732604, from Albany to Chicago, Departure Time: 05:53, Arrival Time: 07:24'
\textcolor{red}{\faClose '2Br Union square Newly furnished' requires staying at least 30 nights, [...]}
} \\
Mind Evolution (ours) &
{\fontfamily{cmss}\selectfont
[Day 1] city: 'from Chicago to Albany', accommodation: 'Mid Town East Side Apartment, Albany', attraction: '-', breakfast: '-', dinner: '-', lunch: '-', transportation': 'Flight Number: F3983350, from Chicago to Albany, Departure Time: 14:29, Arrival Time: 17:03' 
[Day 2] city: 'Albany', accommodation: 'Mid Town East Side Apartment, Albany', attraction: 'Albany Institute of History \& Art, Albany', breakfast: 'Starvin\' Marvin, Albany', lunch: 'Cafe Coffee Day The Square, Albany', dinner: 'PiccoLicko, Albany', 'transportation': '-'
[Day 3] city: 'from Albany to Chicago', accommodation: '-', attraction: 'The McPherson Legacy to the City of Albany - Robert Burns Statue, Albany', breakfast: '-', lunch: '-', dinner: '-', transportation: 'Flight Number: F4008387, from Albany to Chicago, Departure Time: 05:05, Arrival Time: 06:11'
\textcolor{green}{\faCheck}
} \\

    \hline         
    \end{tabular}
    \caption{An example TravelPlanner task and the solutions proposed by \methodname\ and the baselines method.}
    \label{tab:travel_planner_example}
\end{table*}

\begin{table*}[t]
    \setlength{\tabcolsep}{0.4em}
    \small
    \centering
    \arrayrulecolor{gray!70}
    \begin{tabular}{|lp{0.75\textwidth}|}
    \hline
    \multicolumn{2}{|p{0.95\textwidth}|}{
{\fontfamily{cmss}\selectfont
Q: You are visiting San Francisco for the day and want to meet as many friends as possible. Solve the problem by considering various different schedules and picking the best one to optimize your goals.

\vspace{+0.3em}

Travel distances (in minutes):

The Castro to Sunset District: 17. The Castro to Presidio: 20. The Castro to Bayview: 19. The Castro to Chinatown: 20. The Castro to Mission District: 7. Sunset District to The Castro: 17. Sunset District to Presidio: 16. Sunset District to Bayview: 22. Sunset District to Chinatown: 30. Sunset District to Mission District: 24. Presidio to The Castro: 21. Presidio to Sunset District: 15. Presidio to Bayview: 31. Presidio to Chinatown: 21. Presidio to Mission District: 26. Bayview to The Castro: 20. Bayview to Sunset District: 23. Bayview to Presidio: 31. Bayview to Chinatown: 18. Bayview to Mission District: 13. Chinatown to The Castro: 22. Chinatown to Sunset District: 29. Chinatown to Presidio: 19. Chinatown to Bayview: 22. Chinatown to Mission District: 18. Mission District to The Castro: 7. Mission District to Sunset District: 24. Mission District to Presidio: 25. Mission District to Bayview: 15. Mission District to Chinatown: 16.

\vspace{+0.3em}

CONSTRAINTS: 

You arrive at The Castro at 9:00AM. Michelle will be at Sunset District from 6:30PM to 8:30PM. You'd like to meet Michelle for a minimum of 120 minutes. Amanda will be at Presidio from 9:30PM to 10:00PM. You'd like to meet Amanda for a minimum of 30 minutes. Sandra will be at Bayview from 10:00AM to 2:30PM. You'd like to meet Sandra for a minimum of 90 minutes. Kevin will be at Chinatown from 6:15PM to 7:15PM. You'd like to meet Kevin for a minimum of 45 minutes. Mark will be at Mission District from 12:30PM to 1:45PM. You'd like to meet Mark for a minimum of 75 minutes.

}} \\
& \\
\textbf{Method} & \textbf{Answer} \\
1-Pass &
{\fontfamily{cmss}\selectfont
['You start at The Castro at 9:00AM', 'You travel to Bayview in 19 minutes and arrive at 9:19AM', 'You meet Sandra for 90 minutes from 10:00AM to 11:30AM', 'You travel to Mission District in 13 minutes and arrive at 11:43AM', 'You meet Mark for 75 minutes from 12:30PM to 1:45PM', 'You travel to Sunset District in 24 minutes and arrive at 2:09PM', 'You wait until 6:30PM', 'You meet Michelle for 120 minutes from 6:30PM to 8:30PM', 'You travel to Chinatown in 30 minutes and arrive at 8:30PM', 'You wait until 6:15PM', 'You meet Kevin for 45 minutes from 6:15PM to 7:00PM', 'You travel to Presidio in 19 minutes and arrive at 7:19PM', 'You wait until 9:30PM', 'You meet Amanda for 30 minutes from 9:30PM to 10:00PM']
\textcolor{red}{\faClose 'Meeting Mark for 75 minutes from 12:30PM to 1:45PM' doesn\'t match the schedule of Mark, who will be at Mission District from 12:30PM to 01:45PM \textbf{[...]}}
} \\
Best-of-N &
{\fontfamily{cmss}\selectfont
['You start at The Castro at 9:00AM', 'You travel to Bayview in 19 minutes and arrive at 9:19AM', 'You wait until 10:00AM', 'You meet Sandra for 90 minutes from 10:00AM to 11:30AM', 'You travel to Mission District in 13 minutes and arrive at 11:43AM', 'You wait until 12:30PM', 'You meet Mark for 75 minutes from 12:30PM to 1:45PM', 'You travel to Sunset District in 24 minutes and arrive at 2:09PM', 'You wait until 6:30PM', 'You meet Michelle for 120 minutes from 6:30PM to 8:30PM', 'You travel to Chinatown in 30 minutes and arrive at 8:30PM ', 'You wait until 6:15PM', 'You meet Kevin for 45 minutes from 6:15PM to 7:00PM', 'You travel to Presidio in 19 minutes and arrive at 7:19PM', 'You wait until 9:30PM', 'You meet Amanda for 30 minutes from 9:30PM to 10:00PM']
\textcolor{red}{\faClose 'Waiting until 6:15PM' is invalid because the previous step ends after 06:15PM.}
} \\
Sequential Revisions+ &
{\fontfamily{cmss}\selectfont
['You start at The Castro at 9:00AM', 'You travel to Mission District in 7 minutes and arrive at 9:07AM', 'You wait until 12:30PM', 'You meet Mark for 30 minutes from 12:30PM to 1:00PM', 'You travel to Sunset District in 17 minutes and arrive at 1:17PM', 'You wait until 6:30PM', 'You meet Michelle for 30 minutes from 6:30PM to 7:00PM', 'You travel to Presidio in 16 minutes and arrive at 7:16PM', 'You wait until 9:30PM', 'You meet Amanda for 30 minutes from 9:30PM to 10:00PM']
\textcolor{red}{Not meeting with Kevin and Sandra.}
} \\
Mind Evolution (ours) &
{\fontfamily{cmss}\selectfont
['You start at The Castro at 9:00AM', 'You travel to Bayview in 19 minutes and arrive at 9:19AM', 'You wait until 10:00AM', 'You meet Sandra for 90 minutes from 10:00AM to 11:30AM', 'You travel to Mission District in 13 minutes and arrive at 11:43AM', 'You wait until 12:30PM', 'You meet Mark for 75 minutes from 12:30PM to 1:45PM', 'You travel to Chinatown in 16 minutes and arrive at 2:01PM', 'You wait until 6:15PM', 'You meet Kevin for 45 minutes from 6:15PM to 7:00PM', 'You travel to Presidio in 19 minutes and arrive at 7:19PM', 'You wait until 9:30PM', 'You meet Amanda for 30 minutes from 9:30PM to 10:00PM']
\textcolor{green}{\faCheck Not meeting with Michelle, but this is a best possible plan.}
} \\

    \hline         
    \end{tabular}
    \caption{An example Meeting Planning task and the solutions proposed by \methodname\ and the baselines method.}
    \label{tab:meeting_planning_example}
\end{table*}

\clearpage

\ifsteg
\section{Additional Details for \steg}
\label{sec:additional_steg}

The prompt design used for \steg\ is given in \Cref{fig:supp_steg_prompt}.

\paragraph{\steg\ Evaluation}
Each proposed solution should contain a cipher and text component.
The first step is to calculate what is encoded in the text by finding all
the cipher strings;
this is done via simple capitalization-agnostic character-matches.
We refer to the actual encoded string as $M'$.
If $M = M'$ the problem is solved correctly.
The numeric evaluation of a proposed solution is computed as follows:
\begin{enumerate}[leftmargin=1em]
\item Invalid if the text or cipher component cannot be parsed or violates constraints.
    \begin{enumerate}
        \item Words in the cipher cannot be subsets of each other (e.g., origin and original).
        \item Words in the cipher cannot be repeated.
        \item Words in the cipher should be at least 4 characters long.
        \item Words in the cipher should contain only alphabetic characters.
    \end{enumerate}
    \item What is the first position, $i$, in which  $M_i \neq M'_i$?
        This is the integer part of the score.
    \item Compute the Levenshtein distance between $M$ and $M'$.
Levenshtein distance is often used in information theory and linguistics to
measure the difference between two sequences \citep{levDist}.
This is scaled between (0,1) and added to the integer component above.
\end{enumerate}

Additionally, textual feedback, without numeric penalties, is also provided in
the revision request made to the LLM. 
\begin{enumerate}[leftmargin=1em]
    \item A clearly marked list of what $M'$ was found.
    \item A list of number mappings missing from the cipher, or unnecessary numbers specified in the cipher.
    \item If a word appears an incorrect number of times (too few or too many) in the text, it is indicated, along with the error.
    \item An annotated copy of the text is returned.
        The annotations indicate where the cipher-keywords were found (they are shown asterisked), and the first error is indicated. 
    \item If the text encodes the cipher correctly, but also encodes extra words, that is indicated.
    \item If everything in $M'$ is correct, but $|M'| < |M|$, it is indicated as such. 
\end{enumerate}

For this task, we experimented with many different genre forms
(poetry, short-story fiction, essay, monologue, etc.),
as well as inspirations from contemporary to classic writers.

\begin{figure*}[htb]
\tiny
\begin{verbatim}
For this task, you are the world's best poet, linguist and hidden code creator!
You strive to write in the style of shel silverstein.
I would like you to come up with a 1:1 mapping from numbers to words for the list of numbers
demarcated by <HIDDEN-MESSAGE START> and <HIDDEN-MESSAGE END>.
Then, cleverly hide the numbers in an poem about the fun of walking.

SPECIFICS:
* You can choose the words.
* A number cannot have more than one word assigned to it.
* A word cannot be used twice in the mapping.
* Create a substitution-ENCODING-CIPHER for each number and then put 
those words into the poem in the order that they appear in the hidden message.
* If you use a word in the ENCODING-CIPHER in the poem, then it can only be used 
to encode a number.  Nowhere else.  Otherwise it will be mistaken as part of the message.
* The numbers must be encoded in the order that they appear in the hidden message.

<HIDDEN-MESSAGE START>
90 70 50 40 40 10 50 80 70 900 50 30
<HIDDEN-MESSAGE END>

Here are some guidelines to follow.
* The poem must NOT contain any numbers.
* Numbers should not be encoded with a word or words that directly represent the number.  
For example, DO NOT represent "2" : "Dual".  Similarly, "4" : "four" is not acceptable.
* In general, numbers should not be represented by words that are numeric in nature.
* Numbers should not be represented by hyphenated words or using any
other non alphabetic character.  They should be encoded with single words.
* The words used for ciphers should not be subsets of each other.
* Your poem should make sense, be understandable, and be in the style of shel silverstein.
* Your poem should not just be a list of words that look like jibberish.  
Try to have at least 4 words between your cipher words.

Absolutely crucial:
* Important — make sure that each number in the hidden message is represented in the poem.
* DO NOT SKIP ANY numbers. The numbers must be encoded in the SAME order as they are given to you.
* PRECISION IS OF THE UTMOST IMPORTANCE.
* Note that we did not use common words such as 'and' and 'the' etc. to encode the numbers, 
as such words may be difficult to use only at the specific times 
required to encode the hidden-message.
* You should  not use common words to encode the numbers in the cipher.
* The words you use in the cipher should be at least 4 characters long.
* Though you may have to use the same word multiple times, try to avoid using 
the same sentence or phrase multiple times.

Please begin and end your poem with <POEM START> and <POEM END>.

FORMATTING INSTRUCTIONS:
* Please begin your answer by providing the mapping from the numbers in the hidden message 
to the words you select.
* I've included 3 examples of successful ENCODING-CIPHERs below.
* VERY, VERY IMPORTANT:  You should come up with your own creative ENCODING-CIPHERs!
* Note the semicolon that is required after each cipher entry in the ENCODING-CIPHER section.
This must be present!

EXAMPLE #1:

Can you please hide the message demarcated by <HIDDEN-MESSAGE START> and <HIDDEN-MESSAGE END>
into a poem about computers.

<HIDDEN-MESSAGE START>
77 22 33 40 44 77 50 66 55 5 40 40 3 70 8
<HIDDEN-MESSAGE END>

<ENCODING-CIPHER START>
"22" : "computers";
"33" : "become";
"44" : "vital";
"55" : "them";
"66" : "need";
"77" : "everyday";
"40" : "more";
"50" : "need";
"70" : "certain";
"3" : "grow";
"5" : "exist";
"8" : "future";
<ENCODING-CIPHER END>

<POEM START>
Everyday, computers become more vital to our lives.
Everyday, we need them to exist more and more.
That will grow, for certain, in the future.
<POEM END>
\end{verbatim}
\caption{%
An example initial prompt for \steg.  Only 1 of 3 examples is shown.
} 
\label{fig:supp_steg_prompt}
\end{figure*}

\clearpage
\fi

\end{document}